\documentclass{article}
\usepackage{iclr2027_conference,times}

\usepackage{amsmath,amsfonts,bm}

\def\eqref#1{equation~\ref{#1}}

\def\1{\bm{1}}

\DeclareMathAlphabet{\mathsfit}{\encodingdefault}{\sfdefault}{m}{sl}
\SetMathAlphabet{\mathsfit}{bold}{\encodingdefault}{\sfdefault}{bx}{n}

\DeclareMathOperator*{\argmax}{arg\,max}

\usepackage{float}
\usepackage{hyperref}
\usepackage{url}
\usepackage{graphicx}
\usepackage{booktabs}
\usepackage{multirow}
\usepackage{xspace}
\usepackage{adjustbox}
\newcommand{\MethodName}{MKD2World}
\usepackage{placeins}
\usepackage{enumitem}
\usepackage{graphicx}
\usepackage{booktabs}
\usepackage{wrapfig}
\usepackage{caption}
\usepackage{placeins}
\usepackage{booktabs}
\usepackage{amssymb}
\usepackage{makecell}
\usepackage{multirow}
\usepackage{listings}
\usepackage[table]{xcolor}
\usepackage{xcolor}
\usepackage{tcolorbox}
\usepackage{longtable,array}
\tcbuselibrary{listings,breakable}
\lstdefinestyle{prompt}{
  basicstyle=\footnotesize\ttfamily,
  breaklines=true,
  breakatwhitespace=true,
  breakindent=0pt,
  columns=fullflexible,
  keepspaces=true,
  showstringspaces=false
}
\makeatletter
\renewcommand{\hyper@natlinkbreak}[2]{#1}
\makeatother

\title{\MethodName{}: A Typed Spatial Markdown World Model for Web Agents}

\author{
Zhihao Nan$^{1,2,*}$,~
Sifan Zhou$^{3,*}$,~
Yiming Cheng$^{4}$,~
Jinkai Zheng$^{5}$ \\
{\bfseries
Zheng Zhu$^{6}$,~
Si Shi$^{1,\dagger}$,~
Ming Li$^{7,\dagger}$,~
Fei Yu$^{1}$ 
}\\[0.6em]
{\footnotesize
$^{1}$Guangming Laboratory \quad
$^{2}$Shenzhen University \quad
$^{3}$Southeast University \quad
$^{4}$Tsinghua University} \\
{\footnotesize
$^{5}$Hangzhou Dianzi University \quad
$^{6}$GigaAI \quad
$^{7}$The Chinese University of Hong Kong, Shenzhen} \\[0.3em]
{\footnotesize
$^{*}$Equal contribution. \quad
$^{\dagger}$Corresponding authors.
}
}

\iclrfinalcopy
\begin{document}
\maketitle
\thispagestyle{plain}
\pagestyle{plain}

\renewcommand{\headrulewidth}{0pt}
\begin{abstract}
Graphical user interface (GUI) agents can improve decision-making with world-model foresight, 
provided that predicted states accurately reflect post-action interfaces. 
Existing state representations fail to jointly capture interface layout, UI semantics, and explicit locations.
We introduce \textbf{Typed Spatial Markdown (TSM)}, a Markdown-style textual state representation for desktop web interfaces. 
TSM serializes block-level and atomic-level UI elements according to their spatial relationships, 
jointly encoding their types, visual semantics, explicit locations, optional control states, and hierarchical membership. 
To train a screenshot-to-TSM renderer, we construct \textbf{AtomBlock-WebUI}, a dataset with precise annotations at both granularities. 
By injecting element-type labels into HTML and directly extracting bounding boxes after rendering, we avoid annotation errors caused by post-hoc inference from webpage structures.
Building on TSM, we propose \textbf{\MethodName{}}, a desktop web world model trained in two stages 
for current-state description and action-conditioned next-state prediction. 
During inference, the web agent queries \MethodName{} as needed and performs a ``digest'' step, 
in which it interprets each predicted state and extracts task-relevant evidence for selecting the best action.
Experiments show that \MethodName-9B achieves an EleF1 score of 26.50 on the out-of-domain test set, outperforming GPT-5.6-Sol(14.98) by 11.52 points.
Across Online-Mind2Web tasks, \MethodName-9B improves the success rate of the Qwen3.5-35B web agent from 22.94\% to 37.61\%, 
achieving competitive performance among leading open-source GUI world models. Code is available at \url{https://github.com/OmniAI-Lab-CUHKSZ/MKD2World}.
\end{abstract}

\vspace{-3mm}
\section{Introduction}
\vspace{-1mm}
With the development of vision-language models (VLMs), graphical user interface (GUI) agents have become capable of completing interactive tasks in real-world scenarios~\citep{cao2026xiaomigui0technicalreport,hu2026faithfulmobileguiagents,zhou2026qwenuiagenttechnicalreportnextgeneration}. At each decision step, an agent generates an executable action based on the task objective, current interface, and historical information. However, without sufficient understanding of webpage dynamics---including the functions of user interface (UI) controls and the logic of interface transitions---it may generate ineffective actions that increase interaction costs or cause irreversible errors, such as accidentally deleting important data or leaking private information~\citep{shen2026worldmodelaugmentedwebagentsaction,Liu_2026}. Action-conditioned world models offer a promising way to anticipate future interface states before execution, helping agents assess whether candidate actions advance the task while reducing risky real interactions~\citep{mei2026rwom,gu2025webdreamer}.

\begin{figure}[t]
    \centering
    \includegraphics[width=0.95\linewidth]{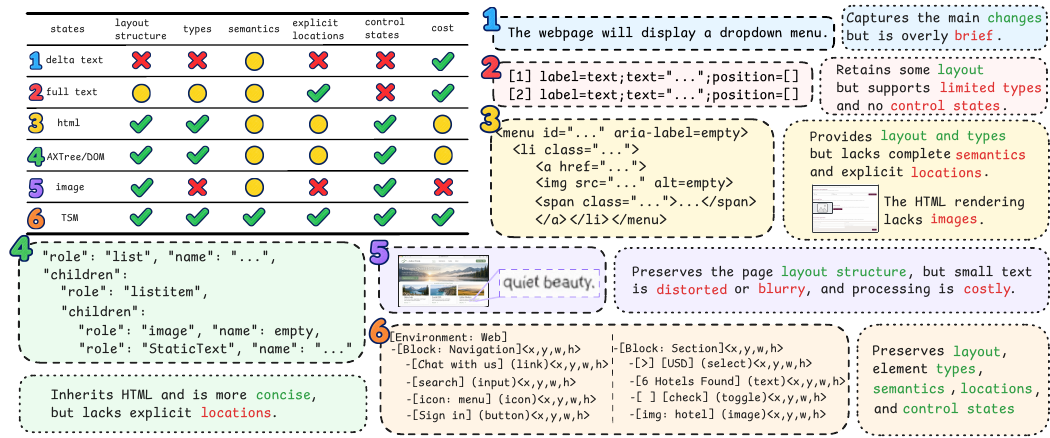}
    \caption{Comparing GUI state representations for world models. TSM better balances a complete, structured interface state with low reconstruction cost.}
    \label{fig:teaser}
    \vspace{-7mm}
\end{figure}

The usefulness of such foresight depends on \emph{what state a world model predicts}. Figure~\ref{fig:teaser} compares representative GUI state representations and highlights the desired trade-off: preserving a complete, structured interface state without incurring the cost of pixel-level reconstruction. We argue that, to distinguish the consequences of candidate actions, a predicted state should preserve a complete yet structured account of the post-action interface, including both the elements that change and the surrounding context needed to interpret those changes---their semantics, locations, control states, and structural relations. At the same time, it should avoid the cost and instability of reconstructing every visual detail of a webpage. Existing GUI world models represent states as free-form descriptions of interface changes, structured interface representations such as the Accessibility Tree (AXTree) and Document Object Model (DOM)~\citep{zuo2026qwenagentworldlanguageworldmodels}, HTML rendered into webpages~\citep{koh2026gworld}, or images generated directly by image generation models~\citep{guan2026cuwm}. Free-form text concisely conveys major changes but loses layout structure and substantial information from the next interface. Common AXTree and DOM serializations preserve parts of the interface structure, yet often omit explicit element locations and have incomplete visual semantics~\citep{wu2023webui}. Other structured textual representations retain some layout and locations, but support limited element types and omit UI control states, complete visual semantics, or interface block structure~\citep{cao2026mobiledreamer}. HTML-rendered images may omit actual images or deviate from the real layout, whereas pure image-generation approaches are costly and often distort fine-grained text~\citep{zheng2026code2world, luo2026vimo}. Thus, these representations do not adequately balance preserving a complete, structured interface state with avoiding the burden of full visual reconstruction. This limits the completeness of the interface information available to a world model and, in turn, to downstream agent reasoning.

To address this bottleneck, we introduce \textbf{Typed Spatial Markdown (TSM)}, a structured and spatially grounded state representation for desktop Web interfaces. TSM organizes rendered Web UIs into block-level and atomic-level elements according to their spatial relationships. Each entry encodes its type, visual semantics, bounding box, optional control state, and hierarchical membership, jointly capturing \emph{what} an element is, \emph{where} it appears, \emph{what} it conveys, and \emph{how} it is organized. Markdown provides a readable hierarchical serialization of this structured state. By retaining the interface information needed to compare action-induced changes without reconstructing the full pixel-level interface, TSM provides a compact and directly comparable state space for action-conditioned world modeling and downstream agent reasoning.

Learning this state space from screenshots requires reliable UI element types and locations. Existing desktop Web UI datasets often extract them from the DOM or AXTree through rule-based parsing~\citep{wu2023webui,gurbuz2026screenparse}. However, HTML structures are highly heterogeneous across webpages: tags and class names are typically tied to page styling and interactions, while role attributes are often incomplete, making unified and reliable semantic annotations difficult to obtain~\citep{gou2025navigatingdigitalworldhumans}. Post-hoc DOM parsing can therefore assign types to visually invisible elements or yield misaligned bounding boxes; visual models may likewise localize elements away from their actual positions~\citep{fu2026guilenscoarsetofinecroppinggui}. To reduce ambiguities introduced by such post-hoc semantic inference, we construct \textbf{AtomBlock-WebUI}. A VLM generates HTML with reference to real webpage layouts and explicitly injects predefined UI element types. After rendering, we directly extract the coordinates of the corresponding typed DOM nodes. This pipeline provides visually aligned type and location supervision for the dual-granularity UI detector in our TSM renderer.

Building on TSM and AtomBlock-WebUI, we introduce \textbf{MKD2World}, an action-conditioned desktop Web world model that predicts future TSM states from screenshots and candidate actions. MKD2World first learns to reconstruct the current interface in the TSM state space, and then learns action-conditioned state transitions with transition-aware objectives that emphasize the elements changed by an action. At inference time, it simulates a next TSM state for each candidate action. We further train the GUI agent to understand TSM and design a digest-based inference strategy, through which the agent interprets predicted typed, spatial, and stateful changes to select the action that best advances the user objective. The main contributions of this work are as follows:
\begin{itemize}[labelindent=0pt, leftmargin=*]
\item \textbf{Typed Spatial Markdown (TSM).} We propose TSM, a structured and spatially grounded Web UI state representation that encodes UI semantics, explicit spatial information, control states, and hierarchical structures, together with a screenshot-to-TSM renderer.
\item \textbf{AtomBlock-WebUI Dataset.} We construct a 9.7k-scale desktop Web UI annotation dataset using a pre-injected automated annotation pipeline, providing element types and bounding boxes for both block-level and atomic-level UI elements to train the dual-granularity UI detector in the screenshot-to-TSM renderer.

\item \textbf{MKD2World: A TSM World Model for Web Agents.} We propose MKD2World, an action-conditioned desktop Web world model that predicts future TSM states from screenshots and candidate actions. MKD2World uses a two-stage training scheme with transition-aware objectives, and supports uncertainty-triggered, digest-based action selection by TSM-aware Web agents.
\item \textbf{Comprehensive Evaluation and Online Validation.} We propose EleF1 to evaluate TSM prediction accuracy, and demonstrate strong out-of-domain prediction and consistent decision-making gains on Online-Mind2Web and VisualWebArena.
\end{itemize}

\vspace{-2mm}
\section{Related Work}
\vspace{-2mm}

\subsection{Web UI Understanding and Datasets}

Existing datasets obtain UI annotations either from webpage structure metadata or directly from screenshots.
Structure-derived datasets---including WebUI~\citep{wu2023webui}, SeeClick~\citep{cheng-etal-2024-seeclick}, and PixelWeb~\citep{yang2025pixelwebwebguidataset}---scale efficiently by leveraging AXTree or DOM information, but their labels depend on heterogeneous HTML, DOM, or accessibility conventions and often require page-specific heuristics to obtain unified semantic types and visually aligned boundaries.
Screenshot-based methods such as OmniParser~\citep{lu2024omniparserpurevisionbased} and ScreenAI~\citep{baechler2024screenaivisionlanguagemodelui} avoid dependence on webpage structure but can suffer from category ambiguity, coordinate offsets, or errors inherited from individual recognition components; high-quality manually annotated datasets~\citep{nayak2025uivisiondesktopcentricguibenchmark,Dicu2025GenGUIAD} are limited in scale.
AtomBlock-WebUI differs from both lines by injecting a predefined semantic taxonomy into webpage HTML before rendering and extracting bounding boxes from the corresponding rendered nodes, providing visually aligned, dual-granularity (block-level and atomic-level) supervision for screenshot-to-TSM rendering.

\subsection{GUI World Models and State Representations}
\vspace{-1mm}

GUI world models support either inference-time action selection~\citep{mei2026rwom,gu2025webdreamer} or training-trajectory synthesis~\citep{xiao2026webworldlargescaleworldmodel,zuo2026qwenagentworldlanguageworldmodels}.
Across both settings, the choice of state representation fundamentally determines which interface information can be predicted, compared, and used for action selection (Figure~\ref{fig:teaser}).
Existing state spaces include: (1)~free-form change descriptions~\citep{chae2025wma,gu2025webdreamer,shen2026wac}, which are compact but lack explicit spatial layouts and hierarchical structure; (2)~structured text or AXTree representations~\citep{cao2026mobiledreamer,zuo2026qwenagentworldlanguageworldmodels,xiao2026webworldlargescaleworldmodel}, which preserve partial structure but do not consistently expose element locations, visually grounded types, and control states in a unified form; (3)~executable HTML code~\citep{zheng2026code2world,koh2026gworld}, which can retain page structure but whose rendered outputs may deviate from the target layout or content; and (4)~future images~\citep{guan2026cuwm,luo2026vimo}, which retain rich visual information but incur high generation cost and are unreliable for fine-grained text and small elements.
These representations typically trade off spatial and hierarchical fidelity against prediction cost and visual grounding.
TSM targets this gap by representing visible UI states as typed, spatially explicit, hierarchically organized elements with control states and visual semantics, remaining efficient for action-conditioned prediction and downstream candidate-action comparison.
AtomBlock-WebUI provides visually aligned supervision for rendering screenshots into this state space, while \MethodName{} models action-conditioned transitions within it for Web-agent reasoning. Detailed discussion refer to Appendix~\ref{detailed_related}.

\vspace{-2mm}
\section{Method}
\label{sec:method}

\vspace{-1mm}
\subsection{Problem Formulation}
\vspace{-1mm}

Given the current screenshot $I_t$ and the task objective $g$, a web agent generates a candidate action set $\mathcal{A}_t$.
For each candidate action $a \in \mathcal{A}_t$, the world model $f_\theta$ predicts a future TSM state $\hat{S}_{t+1}$. Let $V_\psi$ denote a value model that scores predicted TSM states for action selection.
The web agent then selects the best action $a_t^{*}$.
\begin{equation}
\begin{aligned}
\hat{S}_{t+1} = f_\theta(I_t, a),
\qquad
a_t^{*} =
\argmax_{a \in \mathcal{A}_t}
V_\psi\!\left(\hat{S}_{t+1}; g, I_t, a\right).
\end{aligned}
\end{equation}


\vspace{-2mm}
\subsection{Typed Spatial Markdown Representation}

\vspace{-1mm}
\paragraph{Typed Spatial Markdown Definition.} We represent a TSM state as $M_t=(\mathcal{B}_t,\mathcal{E}_t,\rho_t)$, where $\mathcal{B}_t$ and $\mathcal{E}_t$ are the sets of block-level and atomic-level elements, respectively, and $\rho_t$ denotes the hierarchical membership relations among UI elements.
We define 6 block-level types and 8 atomic-level types.
A block-level element is written as $b=(\tau^b,p)$, and an atomic-level element as $e=(u,c,\tau^e,p)$.
Here, $u$ denotes the optional control state, $c$ denotes visible text or visual semantics, and $\tau$ denotes the semantic type.
The bounding box is defined as $p=(x,y,w,h)$, representing the center coordinates, width, and height of a UI element, with values normalized to $[0,999]$.
More details are provided in Appendix~\ref{app:tsm}.

\begin{figure}[t]
    \centering
    \includegraphics[width=0.95\linewidth]{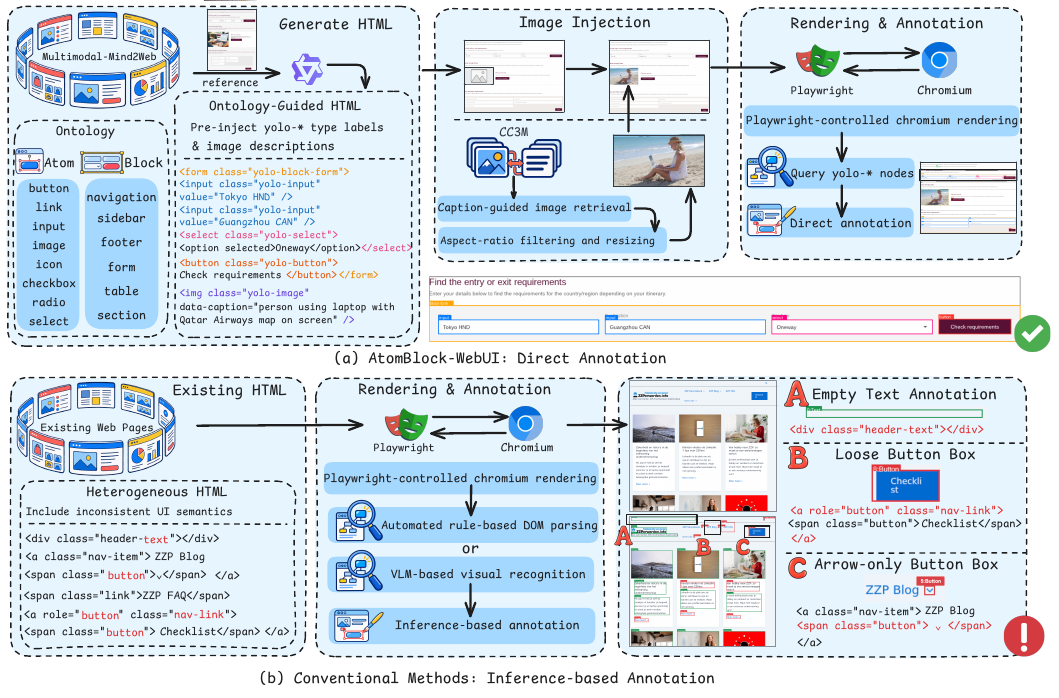}
    \vspace{-3mm}
    \caption{
    Comparison of AtomBlock-WebUI’s direct annotation vs.\ conventional inference-based annotation. 
    (a) AtomBlock-WebUI injects \texttt{yolo-*} labels into HTML, renders in Chromium, and reads boxes straight from those DOM nodes.
    (b) Inference-based annotation (DOM rules or VLMs) on heterogeneous pages yields misaligned boxes.
    }
    \label{fig:atomblock-webui}
    \vspace{-5mm}
\end{figure}

\begin{figure}[h]
\centering

\begin{minipage}{0.70\linewidth}
\centering
\small
\setlength{\tabcolsep}{2pt}

\resizebox{\linewidth}{!}{
\begin{tabular}{lrcccc ccc}
\toprule
\multirow{2}{*}[-0.6ex]{\textbf{Dataset}} &
\multirow{2}{*}[-0.6ex]{\textbf{Image}} &
\multirow{2}{*}[-0.6ex]{\textbf{Annotation Source}} &
\multirow{2}{*}[-0.6ex]{\makecell{\textbf{Viewport}\\\textbf{Controllable}}} &
\multirow{2}{*}[-0.6ex]{\makecell{\textbf{Viewport}\\\textbf{Elements}}} &
\multicolumn{3}{c}{\textbf{Instance share (\%)}}
\\
\cmidrule(lr){6-8}
& & & & &
\textbf{Head$_{20\%}$} &
\textbf{Mid$_{25\%}$} &
\textbf{Tail$_{55\%}$}
\\
\midrule

ScreenParse-v2\citep{gurbuz2026screenparse}
& 1447k
& DOM/AXTree
& $\times$
& 17.7
& 93.33
& 5.91
& 0.76
\\

WebUI-CHI23\citep{wu2023webui}
& 350k
& DOM/AXTree
& $\times$
& 39.1
& 99.05
& 0.84
& 0.11
\\

WebUI-COCO\citep{webui_coco_876}
& 876
& DOM/AXTree
& $\times$
& 33.6
& 63.08
& 29.56
& 7.36
\\

UI-Elements\citep{jain_uielements}
& 544
& DOM/AXTree
& $\times$
& 55.2
& 90.23
& 9.04
& 0.73
\\
\rowcolor{green!10}
AtomBlock-WebUI \textbf{(Ours)}
& 9.7k
& Pre-injected Label
& $\checkmark$
& 48.7
& 61.42
& 30.02
& 8.56
\\

\bottomrule
\end{tabular}
}

\end{minipage}
\hspace{0.015\linewidth}
\begin{minipage}{0.27\linewidth}
\centering
\includegraphics[width=\linewidth]{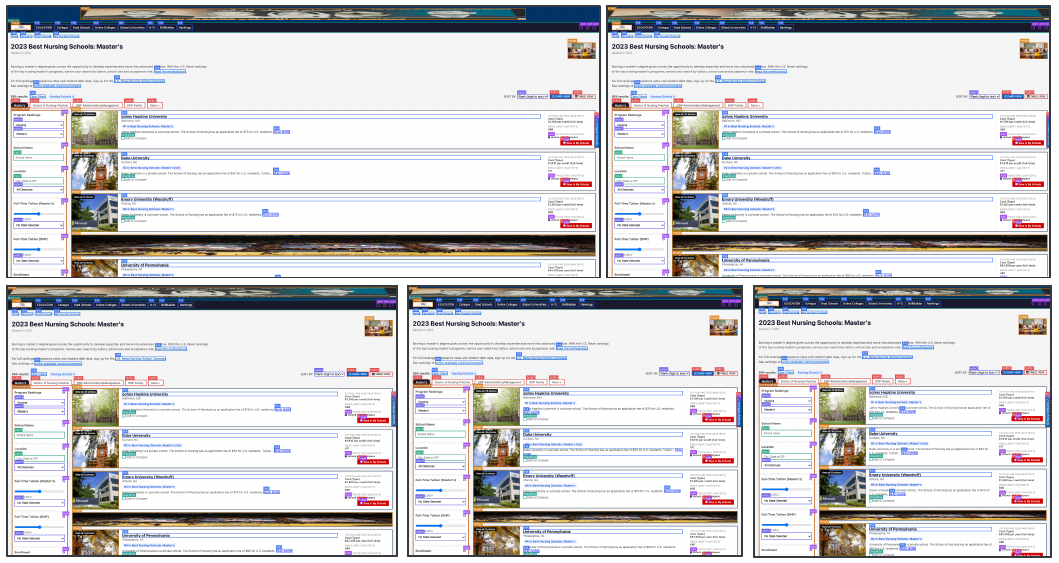}
\end{minipage}
\vspace{-2mm}
\caption{
\textbf{Left}: comparison of WebUI datasets. Head/Mid/Tail report the instance share among types sorted by count, split into the top 20\%, next 25\%, and last 55\%. \textbf{Right}: viewport-controllable renderings of the same page with consistent annotations.
}
\label{fig:ui_datasets}
\vspace{-3mm}
\end{figure}

\vspace{-1mm}
\paragraph{AtomBlock-WebUI Construction.} As shown in Figure~\ref{fig:atomblock-webui}, we first deduplicate the webpage screenshots in Multimodal-Mind2Web \citep{zheng2024gpt4vision} using fingerprint-based filtering. 
We then use the screenshots as layout reference and employ Qwen3.6-Plus to generate HTML with explicit \texttt{yolo-*} labels and image descriptions. 
We retrieve semantically matched images from the CC3M \citep{sharma-etal-2018-conceptual} image--text dataset and inject the image into HTML. 
Finally, we use Playwright to drive Chromium and render full-page screenshots.
The annotation program directly retrieves the \texttt{yolo-*} DOM nodes to
obtain bounding boxes. Figure~\ref{fig:ui_datasets}
compares AtomBlock-WebUI with existing Web UI datasets in annotation source,
viewport controllability, viewport element density, and category balance.
We evaluate annotation quality against existing datasets in
Section~\ref{sec:experiments}. More details are provided in Appendix~\ref{app:atomblock-construction}.


\vspace{-1mm}
\paragraph{Screenshot-to-TSM Renderer.} Given a webpage screenshot, the renderer converts it into TSM using a YOLO26~\citep{jocher2026ultralyticsyolo26} UI detector trained on AtomBlock-WebUI, PP-OCRv6~\citep{zhang2026ppocrv615m345mparameters}, and Qwen3.5-0.8B. 
It first uses the UI detector to identify element types and coordinates and PP-OCRv6 to recognize text, then binds the recognized text to the corresponding elements. 
Some regions are cropped and passed to Qwen3.5-0.8B for visual semantics. 
Finally, the renderer constructs a UI Element Tree with containment relationships and sorts the elements according to their spatial relationships to obtain the TSM representation. Details are in Appendix~\ref{app:tsm-renderer}.


\begin{figure}[t]
    \centering
    \includegraphics[width=0.95\linewidth]{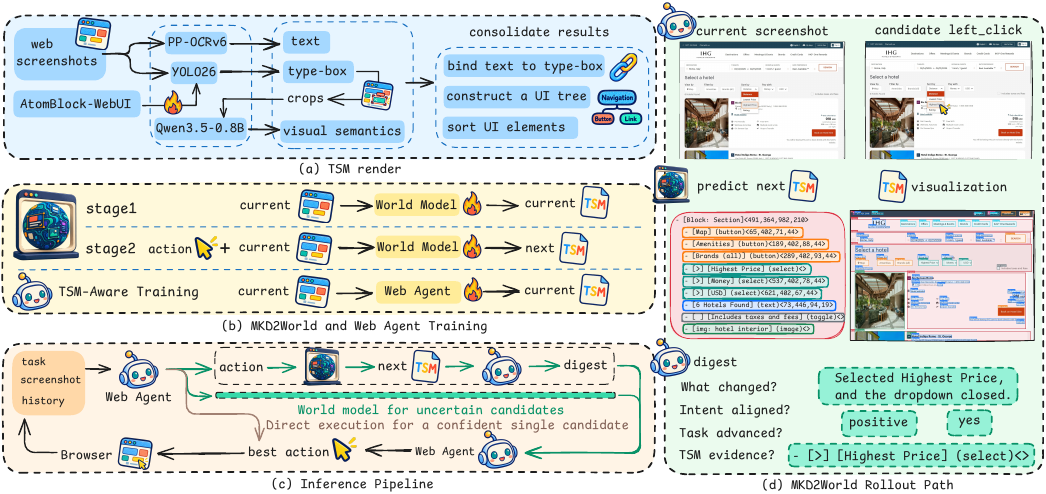}
    \vspace{-2mm}
    \caption{
    \textbf{Overview of \MethodName{}.}
    (a) The TSM renderer combines text, type-box, and visual semantics to convert screenshots into TSM.
    (b) \MethodName{} learns current-state reconstruction and next-state prediction in two stages, followed by TSM-aware web agent training.
    (c) A confident single candidate is executed directly; otherwise, \MethodName{} predicts next states, and the web agent turns them into digests for action selection.
    (d) MKD2World rollout path for one candidate.
    }
    \label{fig:webscene}
    \vspace{-5mm}
\end{figure}

\vspace{-2mm}
\subsection{Action-Conditioned GUI World Model}
\vspace{-1mm}

\paragraph{Dataset Construction.} We extract $(I_t, a_t, I_{t+1})$ from WebChain~\citep{fan2026webchainlargescalehumanannotateddataset} and MolmoWeb-HumanTrajs~\citep{gupta2026molmowebopenvisualweb}, and use the renderer to obtain $(I_t, a_t, I_{t+1}, M_t, M_{t+1})$. 
We then perform quality filtering on the transition pairs and categorize them into \texttt{exact\_noop}, \texttt{near\_noop}, and \texttt{changed} according to the visual differences between the current and next screenshots. 
To optimize the \MethodName{} supervised fine-tuning (SFT) loss, we annotate \texttt{add} and \texttt{update} elements in each transition as focus elements. 
We use the VLM diagnose failure cases and refine
rule-based matching criteria; final focus labels are generated by the
deterministic rules. We manually audit a stratified sample across all action types.  Labels of whether each transition changed and which elements are in focus ultimately supervise \MethodName{} learning. Details are in Appendix~\ref{app:transition-data}.


\paragraph{World Model Training.} We use Qwen3.5-9B as the base model and fine-tune it using LoRA.
Training consists of two stages, as shown in Figure~\ref{fig:webscene}. In the first SFT stage, the model learns to describe the interface, $I_t \rightarrow M_t$, using token-level causal cross entropy.
In the second SFT stage, the model learns to predict the next state, $(I_t,a_t) \rightarrow M_{t+1}$, with the loss for each element $e$ defined as element-level cross entropy:
\begin{equation}
\begin{aligned}
    \mathrm{CE}_{\mathrm{Ele}}(e)
    = -\frac{1}{|e|} \sum_{j \in e}
    \log p_{\theta}\!\left(y_{j} \mid I_t,a_t,y_{<j}\right).
\end{aligned}
\end{equation}
We introduce the Base--Focus Element Loss to emphasize \texttt{add} and
\texttt{update} elements while retaining supervision over the full visible
interface state. For each sample $i$, let $\mathcal{N}_i$ be the set of all elements on the page
and $\mathcal{F}_i \subseteq \mathcal{N}_i$ be the set of focus elements annotated as
\texttt{add} and \texttt{update}.
The loss is
\begin{equation}
    \mathcal{L}_{\mathrm{BF\text{-}Ele}}^{(i)}
    =
    \frac{
        \displaystyle\sum_{e \in \mathcal{N}_i} \mathrm{CE}_{\mathrm{Ele}}^{(i)}(e)
        + \lambda_i \displaystyle\sum_{e \in \mathcal{F}_i} \mathrm{CE}_{\mathrm{Ele}}^{(i)}(e)
    }{
        |\mathcal{N}_i| + \lambda_i |\mathcal{F}_i|
    },
    \qquad
    \lambda_i =
    \begin{cases}
        0, & |\mathcal{F}_i| = 0, \\[4pt]
        \min\!\left(3, \sqrt{\dfrac{|\mathcal{N}_i|}{|\mathcal{F}_i|}}\right),
        & |\mathcal{F}_i| > 0.
    \end{cases}
\end{equation}
where $\lambda_i$ is an adaptive weight that upweights focus elements. When focus elements are sparse, optimizing the average loss over all elements
can underemphasize the localized changes induced by an action. We therefore
scale their contribution according to the ratio
$|\mathcal{N}_i|/|\mathcal{F}_i|$. The square-root scaling provides a
correction, while the cap of $3$ prevents
a small focus set from dominating the page-level objective. When no focus
element is present, $\lambda_i=0$ and the objective reduces to the average
element-level loss over the page.


To further distinguish action-induced transitions from reproducing the
current interface, we introduce a margin-ranking objective
$\mathcal{L}_{\mathrm{margin}}$.
For a transition with a clear visible change, the true next state should be
assigned a lower teacher-forced prediction loss than the current state under
the same conditioning input $(I_t,a_t)$. Let
$\mathrm{CE}_{\mathrm{page}}(M)$ denote the average element-level
cross-entropy incurred when teacher forcing a TSM sequence $M$ conditioned on
$(I_t,a_t)$, and let $\mathcal{C}$ denote the set of \texttt{changed}
transitions. For each sample $i$, we define
\begin{equation}
\begin{aligned}
    \mathcal{L}_{\mathrm{margin}}^{(i)}
    =
    \mathbb{I}[i \in \mathcal{C}]
    \left[
        m
        + \mathrm{CE}_{\mathrm{page}}\!\left(M_{t+1}^{(i)}\right)
        - \mathrm{CE}_{\mathrm{page}}\!\left(M_t^{(i)}\right)
    \right]_{+}.
\end{aligned}
\end{equation}
where $\mathbb{I}[\cdot]$ is the indicator function and $[\cdot]_+$ denotes
the hinge operator. Thus, for a changed transition, the loss is active only
when the model does not prefer the true next state over the current-state
counterfactual by a margin of at least $m$. We apply this objective only to
\texttt{changed} transitions, since treating the current state as a negative
target is not appropriate for \texttt{exact\_noop} or \texttt{near\_noop}
samples. The final training objective is
\begin{equation}
\begin{aligned}
    \mathcal{L}_{\mathrm{final}}
    =
    \frac{1}{B}\sum_{i=1}^{B}
    \left(
        \mathcal{L}_{\mathrm{BF\text{-}Ele}}^{(i)}
        + \beta(s)\,\mathcal{L}_{\mathrm{margin}}^{(i)}
    \right),
\end{aligned}
\end{equation}
where $B$ is the batch size and $\beta(s)$ gradually increases the weight of
$\mathcal{L}_{\mathrm{margin}}$ as the training step $s$ increases.
$\mathcal{L}_{\mathrm{final}}$ encourages \MethodName{} to learn both local
focus elements and global page changes, improving next-state prediction.

\vspace{-1mm}
\paragraph{Evaluation Metrics.} We propose EleF1 to measure the similarity between the TSM representations of two interfaces.
Let $\mathcal{P}$ and $\mathcal{G}$ denote the element sets parsed from the predicted and ground-truth TSM.
For an element pair $(p_i,g_j)$, we define the matching score as
\begin{equation}
\begin{aligned}
    s_{ij}
    = s_{\mathrm{sem}}(p_i,g_j)
    \cdot s_{\mathrm{type}}(p_i,g_j)
    \cdot s_{\mathrm{spa}}(p_i,g_j).
\end{aligned}
\end{equation}
Here, $s_{\mathrm{sem}}$ measures the semantic similarity of element content, $s_{\mathrm{type}}$ measures element-type compatibility, and $s_{\mathrm{spa}}$ measures spatial consistency by jointly considering bounding-box overlap and center distance.
We perform maximum-weight bipartite matching between the two element sets and denote the optimal total matching score by $S^*$.
EleF1 is then defined as
\begin{equation}
\begin{aligned}
    \operatorname{EleF1}(\mathcal{P},\mathcal{G})
    = \frac{2S^*}{|\mathcal{P}|+|\mathcal{G}|}.
\end{aligned}
\end{equation}


\vspace{-4mm}
\subsection{World Model Enhanced Web Agent}
\vspace{-1mm}

At each decision step, the web agent first generates a candidate action set $\mathcal{A}_t=\{(a_k,i_k)\}_{k=1}^{K}$, where $a_k$ is a candidate action and $i_k$ is its intent.
The web agent decides whether to invoke the world model based on its confidence. 
For each candidate action, \MethodName{} takes the current screenshot $I_t$ and the candidate action $a_k$ as input and predicts the interface state $\hat{S}_k$.
The web agent compresses each prediction into a structured digest $d_k=\operatorname{Digest}(I_t,a_k,i_k,\hat{S}_k)$, which includes a textual description of the main changes, their consistency with the action intent, their effectiveness in advancing the task, and evidence from $\hat{S}_k$.
Finally, the web agent aggregates all digests $d_k$ and selects the best action to execute. In addition, we conduct TSM-aware training for the web agent, training it to describe interface states in TSM and improve interaction ability with \MethodName{}.
\vspace{-3mm}
\section{Experiments}
\vspace{-2mm}
\label{sec:experiments}

In this section, we first evaluate the annotation quality of AtomBlock-WebUI. 
We then assess the prediction accuracy of \MethodName{}.
Finally, we enhance web agents with \MethodName{} and evaluate their performance on GUI tasks.

\vspace{-2mm}
\subsection{AtomBlock-WebUI Evaluation}

\vspace{-2mm}
\paragraph{Comparison Datasets.}
We compare AtomBlock-WebUI with four desktop Web UI annotation datasets: ScreenParse-v2, WebUI-CHI23, WebUI-COCO, and UI-Elements. We first deduplicate each dataset.
For ScreenParse-v2 and WebUI-CHI23, we randomly sample 1,000 viewport pages from their training sets.
WebUI-COCO and UI-Elements use all 618 and 508 pages.
AtomBlock-WebUI uses the same set of 1,000 full webpages and reports two evaluation settings:
\texttt{AtomBlock-Top} evaluates only the top 16:9 viewport of each page, while \texttt{AtomBlock-Full} averages the evaluation results over all viewports of the same page. Each dataset uses its native viewport resolution and retains only desktop viewports, predominantly $1920\times1080$.

\vspace{-2mm}
\paragraph{Evaluation.}
Each dataset is annotated under its own category taxonomy.
We use GPT-5.6-Sol with low thinking effort to evaluate annotation quality.
Given the original screenshot, the annotated screenshot, and script-computed statistics comprising drawable-box count in the current viewport, distinct-category count, occupied cells in a \(4\times4\) grid of annotation centers, and HSV color entropy of the original image, the evaluator scores four dimensions: object validity assesses whether recognizable UI elements appear inside the boxes; boundary alignment assesses whether element boundaries align with the annotation boxes; label semantics assesses whether annotated categories are compatible with the element semantics; page complexity is judged from the original image alone and reflects how richly structured the interface is.
We randomly sample 20 pages from each dataset and score them with humans to validate the GPT judge against human ratings. Raters score independently with the same prompt,
blinded to dataset identity and GPT scores (Appendix ~\ref{app:prompts}; Table~\ref{tab:atomblock-quality}).

\begin{table*}[t]
\centering
\caption{Annotation quality comparison among Web UI datasets.}
\label{tab:atomblock-quality}
\small
\setlength{\tabcolsep}{3.5pt}
\resizebox{\textwidth}{!}{%
\begin{tabular}{l cccc cccc}
\toprule
\multirow{2}{*}[-0.6ex]{Dataset}
& \multicolumn{4}{c}{VLM judge}
& \multicolumn{4}{c}{Human} \\
\cmidrule(lr){2-5} \cmidrule(lr){6-9}
& Obj.\ Val.\ $\uparrow$
& Bound.\ Align.\ $\uparrow$
& Label Sem.\ $\uparrow$
& Complexity $\uparrow$
& Obj.\ Val.\ $\uparrow$
& Bound.\ Align.\ $\uparrow$
& Label Sem.\ $\uparrow$
& Complexity $\uparrow$ \\
\midrule
ScreenParse-v2
& 95.93 & 85.87 & 96.75 & 53.39
& 90.00 & 89.25 & 91.25 & 55.75 \\
WebUI-CHI23
& 90.00 & 83.68 & 97.36 & 52.91
& 93.25 & 95.25 & 84.25 & 60.00 \\
WebUI-COCO
& \textbf{99.57} & 93.99 & 84.26 & \underline{71.45}
& 99.25 & \underline{98.00} & 78.25 & 73.00 \\
UI-Elements
& 82.91 & 89.02 & 94.71 & 62.54
& 71.00 & 90.50 & 85.00 & 67.75 \\
\rowcolor{green!10}AtomBlock-Top (Ours)
& \underline{98.82} & \textbf{96.52} & \textbf{99.62} & \textbf{74.14}
& \textbf{100.00} & 97.50 & \underline{94.50} & \textbf{81.00} \\
\rowcolor{green!10}AtomBlock-Full (Ours)
& 98.51 & \underline{95.53} & \underline{99.60} & 68.02
& \underline{99.47} & \textbf{98.37} & \textbf{99.16} & \underline{75.75} \\
\bottomrule
\end{tabular}%
}
\vspace{-5mm}
\end{table*}

\vspace{-2mm}
\paragraph{Annotation Quality.}
AtomBlock-Top leads the VLM judge by 2.53 / 2.26 / 2.69 points on boundary alignment, label semantics, and page complexity, and also ranks top under human evaluation with object validity and the best complexity.
AtomBlock-Full stays close, leading human boundary alignment and label semantics, with only a mild drop in complexity on full pages.

\vspace{-2mm}
\paragraph{Community Adoption.}
AtomBlock-WebUI is released on Hugging Face and has accrued over \textbf{7K} downloads; it also ranked \textbf{Top-5} on the Hugging Face dataset trending leaderboard. These signals suggest that the dataset addresses a clear community need for well-annotated Web UI resources.

\vspace{-2mm}
\subsection{\MethodName{} Prediction Accuracy}
\label{sec:webscene_prediction_accuracy}

\vspace{-2mm}
\paragraph{Datasets and Metrics.}
We extract interface transitions from WebChain and MolmoWeb-HumanTrajs. 
After quality filtering, domain-based splitting, TSM rendering, and focus annotation,
we obtain a 100k Current-State reconstruction training set and a 31k Next-State prediction training set.
We reserve 500 validation transitions for model selection and result analysis,
and evaluate all models on in-domain (ID) and out-of-domain (OOD) test sets.
Each test set contains 1,000 transitions across five action categories:
click, text/key, scroll, navigation, and other, with 200 transitions per category.
We further maintain a balanced website distribution as much as possible~(see Appendix~\ref{app:transition-construction}). For metrics, \textit{EleF1} measures the similarity between the predicted TSM and the ground-truth TSM.
We report \textit{EleF1} for five action categories and use Overall to denote the average value. Format Success measures the ratio of generated TSM outputs that can be successfully parsed.

\vspace{-2mm}
\paragraph{Model Settings.}
\MethodName{} is built on Qwen3.5-9B and trained with the same LoRA configuration in two-stage SFT.
Across both stages, we apply LoRA to all 32 language-model layers and to the attention and MLP modules in each of the 27 vision-encoder blocks,
with rank \(r=32\), \(\alpha=64\), and dropout \(0.05\).
We optimize with AdamW at a learning rate of \(2\times10^{-5}\),
use seed \(0\), cap supervised outputs at \(6{,}000\) tokens, and train on A800 GPUs.
We compare \MethodName{} with four general-purpose multimodal models, including GPT-5.6-Sol, Qwen3.7-Plus, GLM-5V-Turbo, and MiniMax-M3. For all models, we set temperature to \(0\), use low thinking effort, and require predictions in the TSM format.

\begin{table}[t]
    \centering
    \caption{
    Prediction accuracy comparison among world models.
    }
    \label{tab:webscene_prediction_accuracy}
    \small
    \setlength{\tabcolsep}{3pt}
    \resizebox{0.80\linewidth}{!}{
    \begin{tabular}{@{}ll*{9}{c}@{}}
        \toprule
        Split & Model 
        & Click & Text/Key & Scroll & Navigation & Other
        & Overall & Format Success \\
        \midrule
        
        ID
        & GPT-5.6-Sol
        & 16.42 & 19.16 & 12.67 & 7.14 & 17.52
        & 14.58 & \underline{96.70} \\
        
        & Qwen3.7-Plus
        & \underline{19.74} & \underline{22.99} & \underline{12.75} & 7.97 & \underline{21.92}
        & \underline{17.08} & 92.30 \\
        
        & GLM-5V-Turbo
        & 17.59 & 21.19 & 12.11 & \underline{8.06} & 20.63
        & 15.92 & 54.80 \\
        
        & MiniMax-M3
        & 13.31 & 17.41 & 8.86 & 6.78 & 14.60
        & 12.19 & 82.40 \\
        
        \rowcolor{green!10}& \MethodName{} (Ours)
        & \textbf{29.94} & \textbf{38.02} & \textbf{19.40}
        & \textbf{13.41} & \textbf{39.69}
        & \textbf{28.09} & \textbf{97.90} \\
        
        \midrule
        
        OOD
        & GPT-5.6-Sol
        & 15.02 & 15.88 & 12.16 & \underline{11.22} & \underline{20.63}
        & 14.98 & \underline{96.20} \\
        
        & Qwen3.7-Plus
        & \underline{17.47} & \underline{20.90} & \underline{13.10} & 9.06 & 19.18
        & \underline{15.94} & 92.50 \\
        
        & GLM-5V-Turbo
        & 16.43 & 19.73 & 11.97 & 9.02 & 16.59
        & 14.75 & 58.60 \\
        
        & MiniMax-M3
        & 12.66 & 14.53 & 9.30 & 8.10 & 12.53
        & 11.43 & 83.50 \\
        
        \rowcolor{green!10}& \MethodName{} (Ours)
        & \textbf{29.37} & \textbf{39.43} & \textbf{18.02}
        & \textbf{13.80} & \textbf{31.88}
        & \textbf{26.50} & \textbf{98.00} \\
        
        \bottomrule
    \end{tabular}
    }
    \vspace{-6mm}
\end{table}

\vspace{-2mm}
\paragraph{Results and Analysis.}
As shown in Table~\ref{tab:webscene_prediction_accuracy}, 
MKD2World achieves the highest EleF1 on both ID and OOD test sets.
Navigation remains the most challenging category for all methods, likely
because its target interface can change substantially beyond the visible
context of the source page. 
These results demonstrate the strong prediction ability.

\vspace{-2mm}
\subsection{\MethodName{} Enhanced Web Agent}
\vspace{-2mm}

\paragraph{Datasets and Metrics.}
We evaluate the effectiveness of \MethodName{} in improving web agent performance on  Online-Mind2Web \citep{xue2025illusion,deng2023mind2web} and VisualWebArena \citep{koh2024visualwebarena}.
Online-Mind2Web contains 300 real-time website tasks.
After removing unavailable tasks and tasks heavily affected by network instability, we select 109 relatively stable tasks as our test set, including 25 Easy, 48 Medium, and 36 Hard tasks (see Appendix~\ref{app:stable109}).
The interaction trajectories are reviewed by GPT-5.6-Sol under a unified criterion to determine whether each task is successfully completed.
The judge receives the task instruction, the full trajectory of screenshots in temporal order, the action history, the agent's final answer, and runtime error logs.
An early stop by the web agent does not count as success.
We set \(\mathrm{success}=1\) only when the screenshots and action records show that all task requirements are completed;
ambiguous, incomplete evidence, or termination at the \(50\)-step budget yields \(0\).
Additionally, we use 233 tasks from VisualWebArena as a supplementary benchmark to further evaluate the generalization ability of \MethodName{}.
Success on VisualWebArena is determined by the benchmark's official checkers rather than trajectory quality.
After the agent stops, the checkers inspect the final page and the last submitted answer and return a binary score under rules.


\vspace{-2mm}
\paragraph{Model Settings.}
We enhance four web agents with \MethodName{}:
Qwen3.5-35B-A3B, Qwen3.5-9B, GUI-Owl-1.5-8B~\citep{xu2026mobileagentv35}, and deepseek-v4.1-flash. We use WMA, WebDreamer, Qwen-AgentWorld, and WebWorld as world model baselines.
When comparing different world models,
we hold the agent fixed to Qwen3.5-35B-A3B on Online-Mind2Web and to Qwen3.5-9B on VisualWebArena.
All web agents follow the same policy:
the input is the current screenshot, the action history, and the task goal;
the output is only one action type with normalized coordinates.
With world-model support, a web agent may emit one or more candidate actions according to its confidence. We set the context length to 32K tokens, the output length to 5K tokens, and thinking effort to medium.
We cap each episode at 50 steps and stop after 4 consecutive repeated actions.
All world models make predictions under their official prompting strategies.
We cap world-model outputs at 6K tokens, disable thinking mode, and set temperature to \(0\).
For MKD2World, we further fine-tune all web agents except DeepSeek-V4.1-Flash for TSM awareness,
using a 50K subset of the Current-State reconstruction training set at a learning rate of \(2\times10^{-6}\);
all other settings match those used to train MKD2World for Current-State reconstruction in the previous subsection.
We further add a digest step in which the web agent consolidates the MKD2World's predictions before acting.
This adaptation strengthens collaboration between the web agent and MKD2World.

\vspace{-3mm}
\paragraph{Results and Analysis.}
As shown in Tables~\ref{tab:online_mind2web} and~\ref{tab:visualwebarena}, \MethodName{} improves Overall success over the Base agent across most backbones on both benchmarks. Specifically, on Online-Mind2Web with Qwen3.5-35B,
MKD2World reaches \(37.61\%\),
improving over the Base agent by \(14.67\) points and surpassing the strongest prior world model, WebWorld, by \(1.83\) points.
On VisualWebArena with Qwen3.5-9B, MKD2World also achieves the best results.
Under the same condition, world models yield different next-state predictions, and the TSM from \MethodName{} is more concise and complete (see Appendix~\ref{app:comparison-of-worldmodels}).
The gains also transfer across backbones,
suggesting that TSM predictions supply useful evidence for action selection.
How much of this evidence is realized, however, still depends on the agent backbone's own decision making. 


\begin{table}[t]
\centering

\begin{minipage}[t]{0.48\linewidth}
\centering
\caption{Performance on Online-Mind2Web.}
\label{tab:online_mind2web}
\small
\setlength{\tabcolsep}{2pt}
\resizebox{\linewidth}{!}{
\begin{tabular}{lrrrr}
\toprule
& Easy & Medium & Hard & Overall \\
\midrule
\multicolumn{5}{l}{\textbf{Qwen3.5-35B}} \\
\midrule
Base & 44.00 & 22.92 & 8.33 & 22.94 \\
WMA & 64.00 & 33.33 & 16.67 & 34.86 \\
WebDreamer & 40.00 & 35.42 & 13.89 & 29.36 \\
Qwen-AgentWorld & 44.00 & 27.08 & 16.67 & 27.52 \\
WebWorld & 52.00 & 35.42 & 25.00 & 35.78 \\
\rowcolor{green!10}\MethodName{} & 60.00 & 35.42 & 25.00 & 37.61 \\
\midrule
\multicolumn{5}{l}{\textbf{Qwen3.5-9B}} \\
\midrule
Base & 32.00 & 10.42 & 5.56 & 13.76 \\
\rowcolor{green!10}\MethodName{} & 36.00 & 20.83 & 8.33 & 20.18 \\
\midrule
\multicolumn{5}{l}{\textbf{GUI-Owl-1.5-8B}} \\
\midrule
Base & 24.00 & 10.42 & 5.56 & 11.93 \\
\rowcolor{green!10}\MethodName{} & 24.00 & 16.67 & 2.78 & 13.76 \\
\midrule
\multicolumn{5}{l}{\textbf{deepseek-v4.1-flash}} \\
\midrule
Base & 60.00 & 18.75 & 2.78 & 22.94 \\
\rowcolor{green!10} \MethodName{} & 68.00 & 35.42 & 5.56 & 33.03 \\
\bottomrule
\end{tabular}
}
\end{minipage}
\hfill
\begin{minipage}[t]{0.48\linewidth}
\centering
\caption{Performance on VisualWebArena.}
\label{tab:visualwebarena}
\small
\setlength{\tabcolsep}{2pt}
\resizebox{\linewidth}{!}{
\begin{tabular}{lrrrr}
\toprule
& Easy & Medium & Hard & Overall \\
\midrule
\multicolumn{5}{l}{\textbf{Qwen3.5-35B}} \\
\midrule
Base & 37.25 & 19.59 & 12.94 & 21.03 \\
\rowcolor{green!10} \MethodName{} & 49.02 & 19.59 & 16.47 & 24.89 \\
\midrule
\multicolumn{5}{l}{\textbf{Qwen3.5-9B}} \\
\midrule
Base & 31.37 & 11.34 & 9.41 & 15.02 \\
WMA & 27.45 & 8.25 & 5.88 & 11.59 \\
WebDreamer & 39.22 & 13.40 & 4.71 & 15.88 \\
Qwen-AgentWorld & 23.53 & 7.22 & 4.71 & 9.87 \\
WebWorld & 27.45 & 4.12 & 7.06 & 10.30 \\
\rowcolor{green!10} \MethodName{} & 45.10 & 16.49 & 10.59 & 20.60 \\
\midrule
\multicolumn{5}{l}{\textbf{GUI-Owl-1.5-8B}} \\
\midrule
Base & 9.80 & 7.22 & 1.18 & 5.58 \\
\rowcolor{green!10} \MethodName{} & 17.65 & 4.12 & 1.18 & 6.01 \\

\midrule
\multicolumn{5}{l}{\textbf{deepseek-v4.1-flash}} \\
\midrule
Base & 31.37 & 17.53 & 8.24 & 17.17 \\
\rowcolor{green!10} \MethodName{} & 33.33 & 13.40 & 10.59 & 16.74 \\

\bottomrule
\end{tabular}
}
\end{minipage}
\vspace{-4mm}
\end{table}


\vspace{-2mm}
\subsection{Ablation Study}
\label{sec:ablation-study}
\vspace{-2mm}

\begin{wraptable}{r}{0.62\linewidth}
\vspace{-22pt}
\caption{Ablation study of \MethodName{}.}
\label{tab:webscene_task_ablation}
    \vspace{-2mm}
    \begin{adjustbox}{max width=\linewidth}
        \begin{tabular}{cccccccc}

\toprule
\multirow{2}{*}[-0.6ex]{TSM-Aware}
& \multicolumn{2}{c}{\MethodName{}}
& \multirow{2}{*}[-0.6ex]{Digest}
& \multirow{2}{*}[-0.6ex]{Easy}
& \multirow{2}{*}[-0.6ex]{Medium}
& \multirow{2}{*}[-0.6ex]{Hard}
& \multirow{2}{*}[-0.6ex]{Overall} \\
\cmidrule(lr){2-3}
& \scriptsize$\mathcal{L}_{\mathrm{BF\text{-}Ele}}$
& \scriptsize$\mathcal{L}_{\mathrm{margin}}$
& & & & & \\
\midrule
--         & -- & -- & --                         & 44.00 & 22.92 & 8.33 & 22.94 \\
\checkmark & -- & -- & --                         & 32.00 & 12.50 & 2.78 & 13.76 \\
-- & \checkmark & -- & --                         & 56.00 & 25.00 & 16.67 & 29.36 \\
-- & \checkmark & \checkmark & --                 & 56.00 & 29.17 & 19.44 & 32.11 \\
-- & \checkmark & \checkmark & \checkmark         & 56.00 & 37.50 & 19.44 & 35.78 \\
\rowcolor{green!10}\checkmark & \checkmark & \checkmark & \checkmark & 60.00 & 35.42 & 25.00 & 37.61 \\
\bottomrule
\end{tabular}  
    \end{adjustbox}
    \vspace{-4mm}
\end{wraptable}
\textbf{Ablation Study.}
We ablate the components of \MethodName{} on Online-Mind2Web with Qwen3.5-35B as the web agent.
As shown in Table~\ref{tab:webscene_task_ablation},
we vary TSM-aware agent training,
the losses \(\mathcal{L}_{\mathrm{BF\text{-}Ele}}\) and \(\mathcal{L}_{\mathrm{margin}}\),
and the digest stage. Removing all components yields a base agent with \(22.94\%\) overall success.
Adding \(\mathcal{L}_{\mathrm{BF\text{-}Ele}}\) alone raises overall success to \(29.36\%\). Further adding \(\mathcal{L}_{\mathrm{margin}}\) raises it to \(32.11\%\). This shows that the loss design lets the world model learn both locally changing elements and global interface changes, improving prediction accuracy.
Enabling digest then reaches \(35.78\%\),
consistent with letting the agent consolidate MKD2World's predictions before acting.
TSM-aware training alone, however, drops overall success to \(13.76\%\).
This shows that adapting the agent to TSM is not helpful by itself;
without \MethodName{} predictions to condition on,
the adaptation harms decision making.
Combining TSM-aware training with the full world-model stack recovers and further improves performance to \(37.61\%\),
especially on Hard (\(25.00\%\)),
indicating that TSM-aware training strengthens collaboration with \MethodName{}.
\vspace{-3mm}
\section{Conclusion}
\vspace{-2mm}
In this paper, we presented TSM, a complete structured interface state, and constructed viewport-controllable AtomBlock-WebUI with balanced types and precise boxes.
We further introduced \MethodName{} and proposed digest and TSM-awareness to strengthen WebAgent--\MethodName{} collaboration.
Empirically, \MethodName{} outperforms GPT-5.6-Sol on prediction accuracy and boosts Qwen3.5-35B by $+14.67\%$ on Online-Mind2Web.
Despite the improvements, due to the complexity and frequent updates of web interfaces, \MethodName{} cannot achieve pixel-level aligned prediction. Our current setting targets static pages; it does not yet perform temporal understanding of audio, video, or animation, nor predict audiovisual changes over the next time interval.

\subsection*{AI use statement}
In this work, we used generative AI tools to generate part of the dataset
metadata, followed by manual quality checks; to score dataset annotation
quality and Online-Mind2Web task completion, with randomly sampled human
judgments to assess agreement; to generate icons for figures; and to adjust
\LaTeX{} formatting.
We have reviewed all AI-assisted work and take responsibility for the final content of this work.

\subsection*{Reproducibility statement}

We provide additional details in the appendix.
Construction of AtomBlock-WebUI is described in Appendix~\ref{app:atomblock-construction},
TSM renderer processing in Appendix~\ref{app:tsm-renderer},
and EleF1 evaluation in Appendix~\ref{app:elef1-metrics}.
Appendix~\ref{app:transition-construction} details the construction of the MKD2World training and evaluation datasets and reports their distributions.
Focus-element annotation rules for transition samples are given in Appendix~\ref{app:focus-annotation}.
For Online-Mind2Web, after excluding unavailable tasks and those heavily affected by network instability, we evaluate on a relatively stable subset of 109 tasks; the task list is provided in Appendix~\ref{app:stable109}.
The main prompts used in this work are listed in Appendix~\ref{app:prompts}.
Our code is provided in the supplementary materials.

\bibliography{refs}
\bibliographystyle{iclr2027_conference}

\appendix
\clearpage
\appendix
\section{MKD2World Prediction Examples}
\label{app:transition-examples}
\begin{center}
  \includegraphics[width=\textwidth]{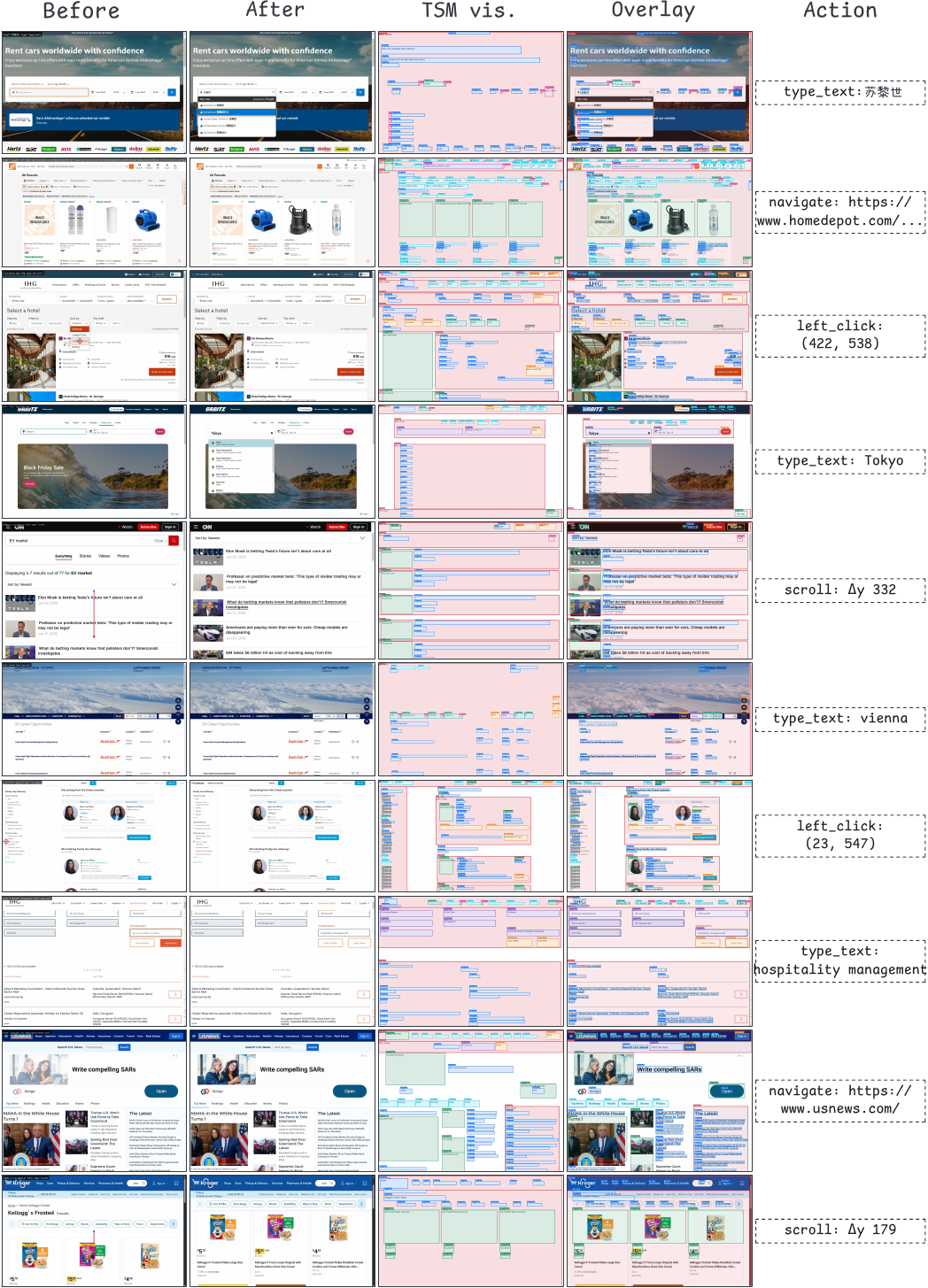}
\end{center}
\captionof{figure}{MKD2World prediction examples. Each row shows the before/after screenshots, the TSM visualization, its overlay on the after screenshot, and the action.}
\label{fig:transition-examples}

\section{Comparison of World Models}
\label{app:comparison-of-worldmodels}
\begin{center}
\includegraphics[width=\textwidth,height=0.42\textheight,keepaspectratio]{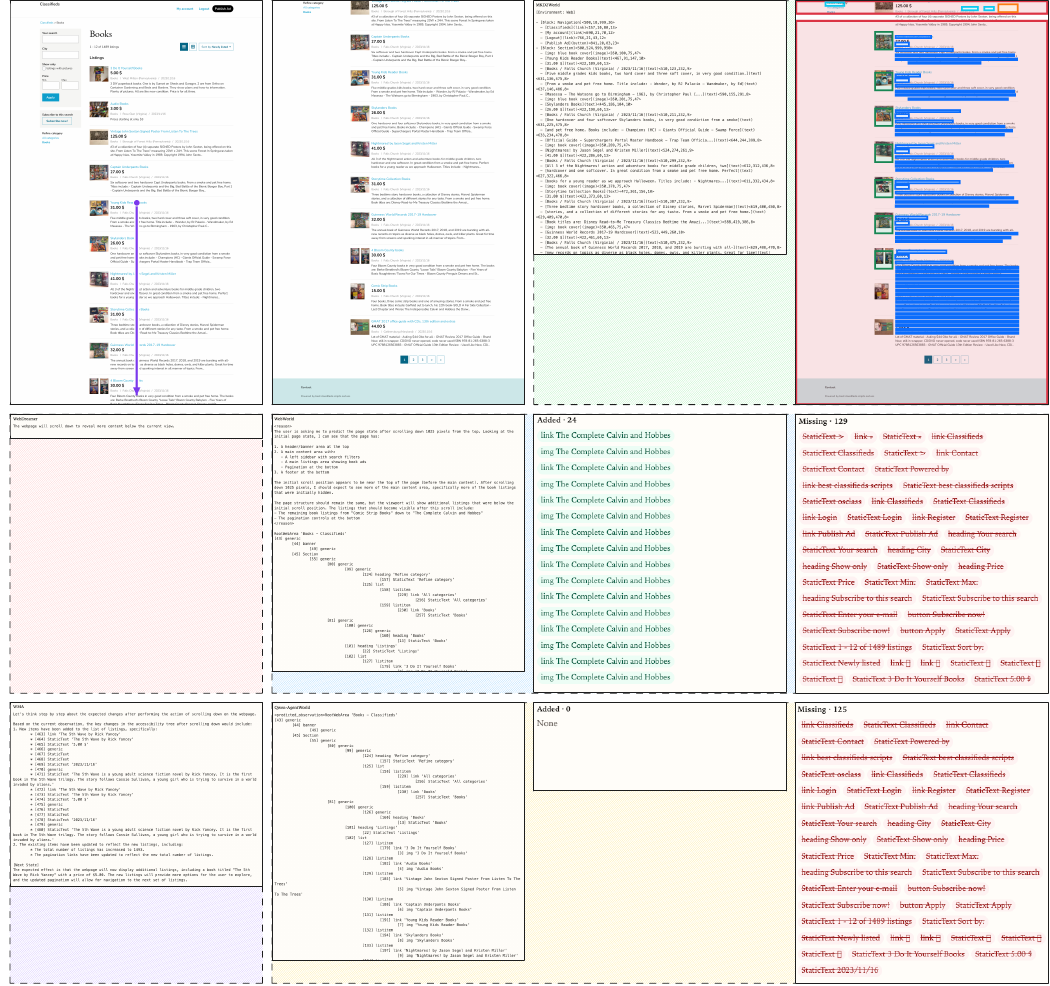}
\end{center}
\captionof{figure}{Comparison of world models under the same observation and action. The action is scrolling down the Books listings on Classifieds.}

\begin{center}
\includegraphics[width=\textwidth,height=0.42\textheight,keepaspectratio]{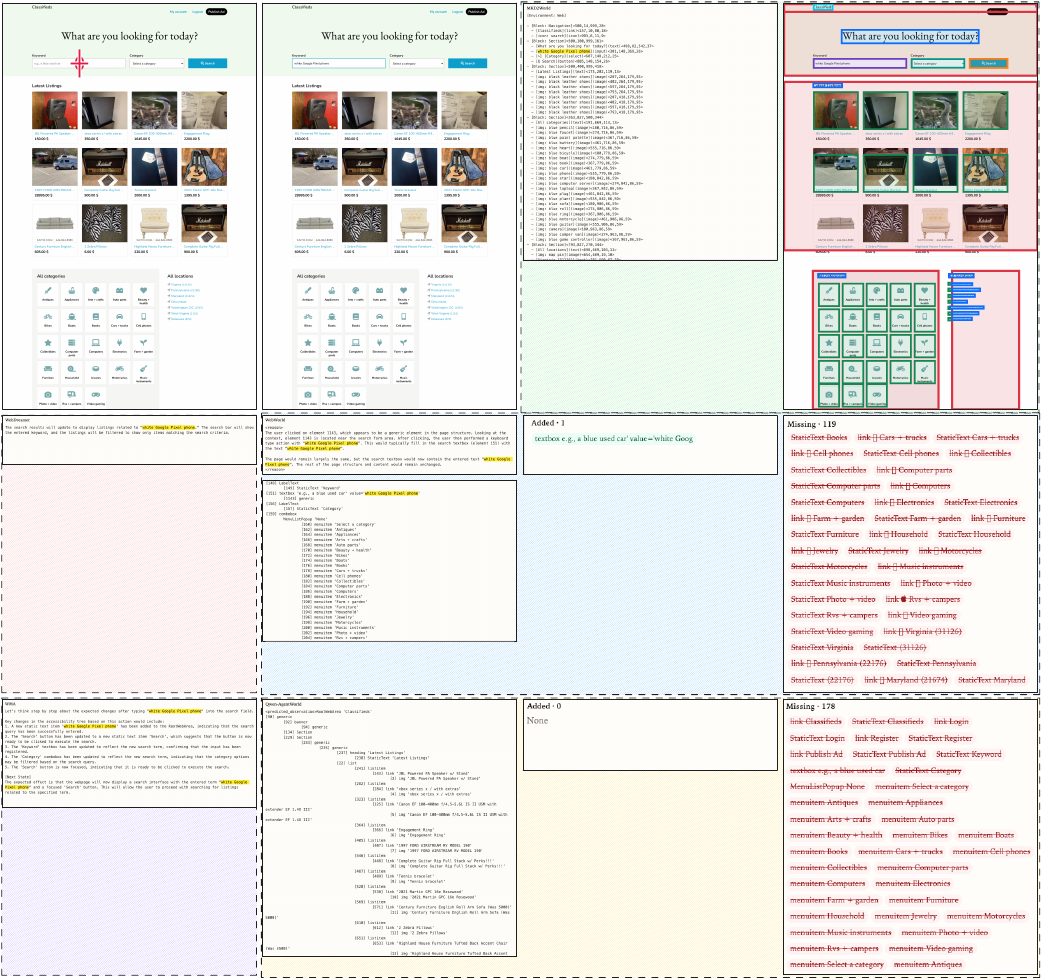}
\end{center}
\captionof{figure}{Comparison of world models under the same observation and action. The action is typing “white Google Pixel phone” into the search box on Classifieds.}

\begin{center}
\includegraphics[width=\textwidth,height=0.42\textheight,keepaspectratio]{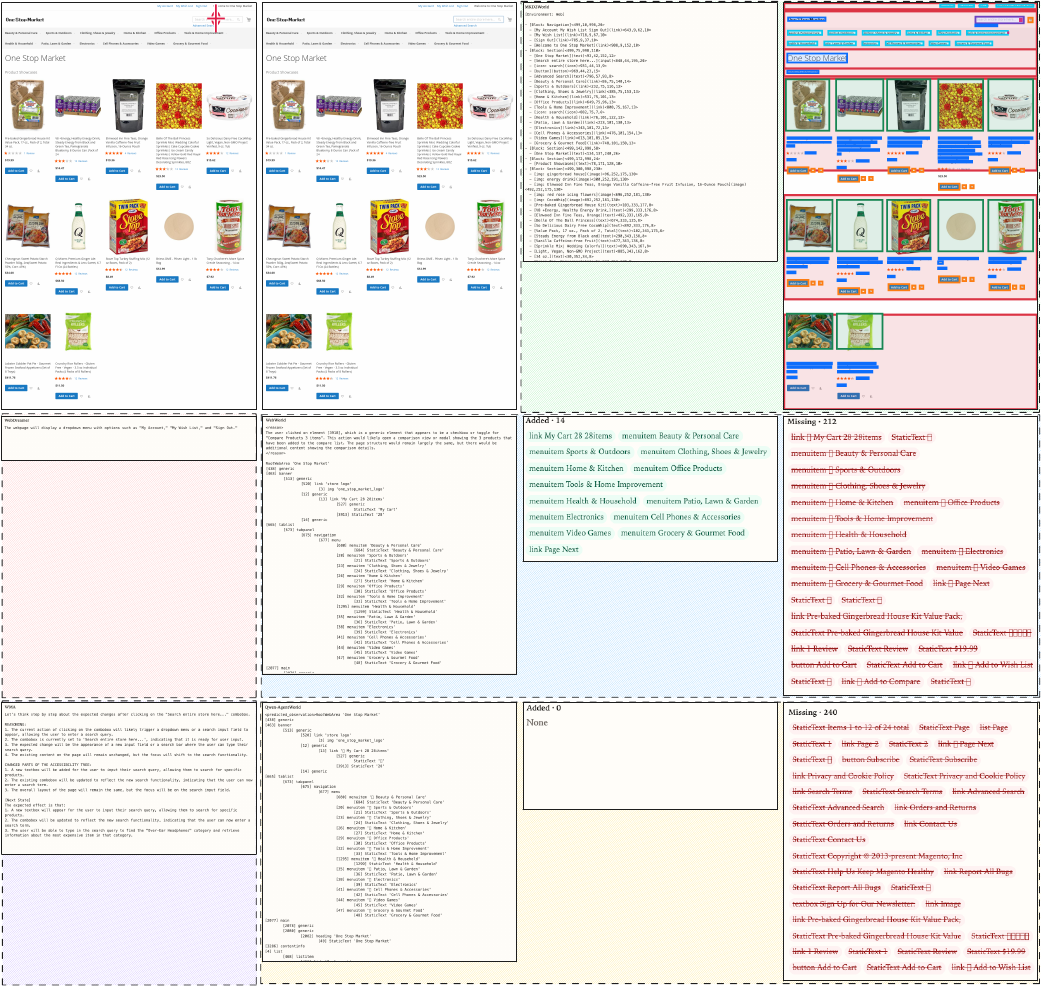}
\end{center}
\captionof{figure}{Comparison of world models under the same observation and action. The action is clicking the search box on the Market homepage.}

\section{AtomBlock-WebUI Construction and Statistics}
\label{app:atomblock-construction}

We perform fingerprint-based near-duplicate filtering on screenshots from Multimodal-Mind2Web. Each image is converted to grayscale, resized to $17 \times 16$ with Lanczos interpolation, and encoded as a 256-bit difference hash (dHash) by comparing adjacent pixels in each row. Screenshots whose hashes have Hamming distance 
$d_H(h_i,h_j)=\operatorname{popcount}(h_i\oplus h_j)\leq5$ are connected in a graph; each connected component keeps one randomly selected image. 
Subsequently, we use the filtered screenshots as layout references and generate HTML with Qwen3.6-Plus. We require the model to cover 8 atomic and 6 block types, inject \texttt{yolo-*} category labels and vary color schemes for visual diversity. Each image node is an \texttt{<img>} tag with a \texttt{data-caption} attribute.
We next inject CC3M images into pages. For each \texttt{<img>} node, we encode its \texttt{data-caption} with MiniLM-L6-v2 and retrieve the top-50 candidates via FAISS. We then re-rank by aspect-ratio mismatch with the target region and randomly pick one of the top-5. The chosen image is resized to the node and written back to the HTML.
We render pages with Playwright-driven Chromium and collect bounding boxes from nodes tagged with \texttt{yolo-*} labels. We retain the original image-injected HTML, so each page can be re-rendered under different viewports while producing consistent annotations. AtomBlock-WebUI comprises 9,683 full-page screenshots and 1,321,234 type--bounding box annotations, split into 6,906, 1,388, and 1,389 pages for training, validation, and testing.
\begin{table}[t]
\caption{Category distribution of AtomBlock-WebUI.}
\label{tab:atomblock_statistics}
\centering
\small
\setlength{\tabcolsep}{8pt}
\begin{tabular}{clcrr}
\toprule
ID & Category & Granularity & Instances & Proportion (\%) \\
\midrule
0  & \texttt{button}        & Atomic & 113,089 & 8.56  \\
1  & \texttt{link}          & Atomic & 626,321 & 47.40 \\
2  & \texttt{input}         & Atomic & 18,520  & 1.40  \\
3  & \texttt{image}         & Atomic & 184,878 & 13.99 \\
4  & \texttt{icon}          & Atomic & 185,215 & 14.02 \\
5  & \texttt{checkbox}      & Atomic & 42,887  & 3.25  \\
6  & \texttt{radio}         & Atomic & 4,431   & 0.34  \\
7  & \texttt{select}        & Atomic & 20,179  & 1.53  \\
\midrule
8  & \texttt{block-nav}     & Block  & 11,424  & 0.86  \\
9  & \texttt{block-sidebar} & Block  & 3,712   & 0.28  \\
10 & \texttt{block-footer}  & Block  & 9,058   & 0.69  \\
11 & \texttt{block-form}    & Block  & 2,109   & 0.16  \\
12 & \texttt{block-table}   & Block  & 697     & 0.05  \\
13 & \texttt{block}         & Block  & 98,714  & 7.47  \\
\midrule
\multicolumn{3}{l}{Total} & 1,321,234 & 100.00 \\
\bottomrule
\end{tabular}
\end{table}
Table~\ref{tab:atomblock_statistics} summarizes AtomBlock-WebUI's category distribution: 8 atomic categories account for 1,195,520 instances(90.49\%), and 6 block categories for 125,714(9.51\%).

\section{Related Work}
\label{detailed_related}

\subsection{Web UI Understanding and Datasets}

Web UI understanding requires identifying visible interface elements together with their semantic types and spatial extents. Existing datasets and methods obtain such annotations primarily from webpage structure metadata or from visual recognition.

\textbf{1. Structure-derived annotations.}
WebUI~\citep{wu2023webui} constructs UI datasets through rule-based parsing of the AXTree, extracting UI elements and their corresponding bounding boxes.
SeeClick~\citep{cheng-etal-2024-seeclick} and MultiUI~\citep{liu2024harnessingwebpageuistextrich} similarly derive annotations from webpage structural metadata, but are commonly used for task-oriented grounding rather than dense full-page annotation of all visible UI elements.
WebCode2M~\citep{Gui_2025} collects real webpage screenshots and DOM-node information to annotate webpage structures, whereas PixelWeb~\citep{yang2025pixelwebwebguidataset} obtains visible element boundaries through DOM information and pixel-level rendering, with its public release including substantial mobile-web data.
These approaches benefit from available webpage metadata, but their labels depend on heterogeneous HTML, DOM, or accessibility structures; obtaining a unified semantic taxonomy and visually aligned element boundaries therefore often requires page-specific heuristics. \textbf{2. Screenshot-based UI understanding.}
Visual methods instead infer UI elements directly from screenshots.
OmniParser~\citep{lu2024omniparserpurevisionbased} detects interactive regions but covers a limited set of categories and can be sensitive to model parameters.
ScreenAI~\citep{baechler2024screenaivisionlanguagemodelui} combines OCR, UI detectors, and image captioning models to generate large-scale annotations, whose quality may inherit errors from individual recognition components.
ScreenParse~\citep{gurbuz2026screenparse} uses VLMs for UI-element category prediction and quality filtering.
UICVD~\citep{nayak2025uivisiondesktopcentricguibenchmark} and GenGUI~\citep{Dicu2025GenGUIAD} rely on manual annotation and provide high-quality labels, but at limited scale.
While screenshot-based approaches avoid dependence on webpage structure, they may still suffer from category ambiguity and coordinate offsets. \textbf{AtomBlock-WebUI} differs from both lines of work by injecting a predefined semantic taxonomy into webpage HTML before rendering and extracting bounding boxes from the corresponding rendered nodes.
It thereby provides visually aligned, dual-granularity annotations over both block-level and atomic-level elements.
Rather than targeting only task-relevant grounding or generic UI detection, it is designed to supervise screenshot-to-structured-state rendering for TSM construction.

\subsection{GUI World Models and State Representations}

GUI world models follow two broad paradigms.
For inference-time decision enhancement, a world model predicts the next interface state conditioned on each candidate action, and a value model scores the predictions to select the best action~\citep{mei2026rwom, gu2025webdreamer}.
For training data synthesis, agents interact with world models over multiple steps to produce interaction trajectories that are then used for agent training~\citep{xiao2026webworldlargescaleworldmodel,fang2025webevolverenhancingwebagent,zuo2026qwenagentworldlanguageworldmodels}.
Across both settings, the choice of state representation fundamentally
determines which interface information can be predicted, compared, and used for action selection. Figure~\ref{fig:teaser} summarizes the central trade-off among representative GUI state representations: preserving rich, structured interface information while avoiding the cost and instability of pixel-level reconstruction. \textbf{1. Free-form textual change descriptions.}
WMA~\citep{chae2025wma}, WebDreamer~\citep{gu2025webdreamer}, and WAC~\citep{shen2026wac} describe post-action interface changes as free-form natural-language text.
This representation is compact and easy for language models to generate, but typically does not explicitly encode element-level spatial layouts, does not provide a structured, comparable state, and makes it difficult to evaluate prediction accuracy or directly compare candidate outcomes. \textbf{2. Structured textual and accessibility representations.}
MobileDreamer~\citep{cao2026mobiledreamer} uses structured text to describe mobile Web interfaces, capturing some hierarchical layout but supporting only a limited set of element types and omitting many interactive elements and block-level structures.
Qwen-AgentWorld~\citep{zuo2026qwenagentworldlanguageworldmodels} and WebWorld~\citep{xiao2026webworldlargescaleworldmodel} predict AXTree representations in Web environments, and more broadly, AXTree and DOM have been widely adopted as structured observations for GUI agents.
However, these representations do not consistently expose explicit element locations or visually grounded semantics in a form suitable for state comparison, hindering direct use for pixel-aligned state prediction and candidate-action evaluation. \textbf{3. Code and image state representations.}
Code2World~\citep{zheng2026code2world} and gWorld~\citep{koh2026gworld} predict future states as HTML code that is subsequently rendered into images; however, the rendered pages often lack realistic visual assets and can diverge from real webpage layouts.
CUWM~\citep{guan2026cuwm} predicts future states directly as interface images, but incurs high generation cost and tends to produce distorted or blurry small text.
ViMo~\citep{luo2026vimo} masks text regions, predicts a text-free image, and uses an LLM to infill text, but this decoupled generation lacks tight semantic grounding in spatial elements and can produce inaccurate fine-grained content.

Prior work has studied UI understanding datasets, GUI element grounding, and action-conditioned world modeling, but these lines have not jointly provided a state space that is spatially explicit, hierarchically structured, and efficient to predict. \textbf{TSM} fills this gap by serializing complete UI states into typed spatial elements. \textbf{AtomBlock-WebUI} supplies aligned supervision for rendering screenshots into this state space, and \textbf{MKD2World} models action-conditioned transitions within it.

\section{TSM}
\label{app:tsm}

\subsection{TSM syntax mapping}
\label{app:tsm-syntax-mapping}

\begin{table}[t]
    
    \centering
    \caption{TSM syntax mapping.}
    \label{tab:tsm_syntax}
    \small
    \begin{tabular}{ll}
        \toprule
        \textbf{UI Element} & \textbf{TSM Representation} \\
        \midrule
        Text              & \texttt{[content](text)<cx,cy,w,h>} \\
        Link              & \texttt{[content](link)<cx,cy,w,h>} \\
        Button            & \texttt{[content](button)<cx,cy,w,h>} \\
        Input             & \texttt{[content](input)<cx,cy,w,h>} \\
        Icon              & \texttt{[icon:semantics](icon)<cx,cy,w,h>} \\
        Image             & \texttt{[img:semantics](image)<cx,cy,w,h>} \\
        Unchecked toggle  & \texttt{[ ] [content](toggle)<cx,cy,w,h>} \\
        Checked toggle    & \texttt{[x] [content](toggle)<cx,cy,w,h>} \\
        Expanded dropdown & \texttt{[v] [content](select)<cx,cy,w,h>} \\
        Collapsed dropdown& \texttt{[>] [content](select)<cx,cy,w,h>} \\
        Semantic block    & \texttt{[Block:semantics]<cx,cy,w,h>} \\
        Environment       & \texttt{[Environment:Web]} \\
        \bottomrule
    \end{tabular}
\end{table}

UI Detector labels are mapped to TSM types by merging \texttt{checkbox}/\texttt{radio} into \texttt{toggle}, renaming \texttt{block-*} to Navigation/Sidebar/Footer/Form/Table, and mapping \texttt{block} to Section. Table~\ref{tab:tsm_syntax} lists the TSM representation of each element type.

\subsection{TSM Renderer Details}
\label{app:tsm-renderer}

The screenshot renderer converts a webpage screenshot into a TSM state using a YOLO26 detector trained on AtomBlock-WebUI, PP-OCRv6, and Qwen3.5-0.8B. A child block is contained in a parent if its center lies inside the parent and the parent covers more than 85\% of its area. The block priorities are 0 for \texttt{Navigation}, \texttt{Sidebar}, and \texttt{Footer}, 1 for \texttt{Form} and \texttt{Table}, and 2 for \texttt{Section} (lower is high priority). So generic \texttt{Section} do not override function-specific blocks. For each atomic element, we first take blocks that contain its center; otherwise those covering at least 50\% of its area. Among candidates, we keep the smallest block with the highest priority; unmatched elements stay at the TSM top level.
Qwen3.5-0.8B describes images and icons, and judges states of checkboxes, radios, and selects. For checkboxes and radios, we enlarge the crop to capture the same line. For selects, we extend the crop downward and treat the dropdown as open if options appear.

\section{EleF1 Metrics}

\label{app:elef1-metrics}

EleF1 measures element-level agreement between the predicted and the ground-truth states. Let $\mathcal{P}$ and $\mathcal{G}$ be the element sets of the predicted and ground-truth TSM. 
For $p_i \in \mathcal{P}$ and $g_j \in \mathcal{G}$, the matching score is
$$
s_{ij}
=
s_{\mathrm{sem}}(p_i,g_j)
\cdot
s_{\mathrm{type}}(p_i,g_j)
\cdot
s_{\mathrm{spa}}(p_i,g_j),
$$
where $s_{\mathrm{sem}}$, $s_{\mathrm{type}}$, and $s_{\mathrm{spa}}$ measure semantic, type and spatial consistency.
We embed the element content $t_i^p$ and $t_j^g$ with \texttt{MiniLM-L12-v2} and compute their cosine similarity
$$
c_{ij}
=
\cos\!\left(
\operatorname{Emb}(t_i^p),
\operatorname{Emb}(t_j^g)
\right),
$$
which is mapped to a semantic score by a piecewise-linear function:
$$
s_{\mathrm{sem}}(p_i,g_j)
=
\begin{cases}
0, & c_{ij}<0.55,\\[4pt]
0.5\cdot\dfrac{c_{ij}-0.55}{0.65-0.55},
& 0.55\le c_{ij}<0.65,\\[10pt]
0.5+0.5\cdot\dfrac{c_{ij}-0.65}{1-0.65},
& c_{ij}\ge 0.65.
\end{cases}
$$
Type compatibility for atomic types is given in Table~\ref{tab:type_compatibility}.
Every semantic block is parsed as type \texttt{block}:
$s_{\mathrm{type}}=1$ when both elements are blocks, and
$s_{\mathrm{type}}=0.5$ when exactly one is a block.
Spatial consistency combines IoU and center distance:
\begin{table}[t]
\caption{Type compatibility matrix for atomic elements.}
\label{tab:type_compatibility}
\centering
\small
\setlength{\tabcolsep}{4pt}
\begin{tabular}{lcccccccc}
\toprule
& text
& link
& button
& image
& icon
& toggle
& select
& input \\
\midrule
text   & 1.0 & 0.5 & 0   & 0   & 0   & 0   & 0   & 0   \\
link  & 0.5 & 1.0 & 0.5 & 0.5 & 0.5 & 0   & 0   & 0   \\
button & 0   & 0.5 & 1.0 & 0.5 & 0.5 & 0   & 0   & 0   \\
image   & 0   & 0.5 & 0.5 & 1.0 & 0.5 & 0   & 0   & 0   \\
icon  & 0   & 0.5 & 0.5 & 0.5 & 1.0 & 0   & 0   & 0   \\
toggle & 0   & 0   & 0   & 0   & 0   & 1.0 & 0   & 0   \\
select & 0   & 0   & 0   & 0   & 0   & 0   & 1.0 & 0   \\
input  & 0   & 0   & 0   & 0   & 0   & 0   & 0   & 1.0 \\
\bottomrule
\end{tabular}
\end{table}

\[
s_{\mathrm{spa}}(p_i,g_j)
=
\alpha\sqrt{\operatorname{IoU}(p_i,g_j)}
+
(1-\alpha)
\exp\!\left(
-\frac{d_{ij}^{2}}{2\sigma^{2}}
\right),
\]
where $d_{ij}$ is the normalized center distance, $\alpha=0.5$, and $\sigma=0.05$. The distance term avoids over-penalizing small, nearby elements with non-overlapping boxes.

With $\Omega$ a one-to-one matching obtained by maximum-weight bipartite matching,
\[
\operatorname{EleF1}(\mathcal{P},\mathcal{G})
=
\frac{2}{|\mathcal{P}|+|\mathcal{G}|}
\max_{\Omega}
\sum_{(i,j)\in\Omega} s_{ij}.
\]
The normalization penalizes both missing and redundant elements.

\section{Transition Dataset and Focus Annotation}
\label{app:transition-data}

\subsection{Transition Dataset Construction and Statistics}
\label{app:transition-construction}

\begin{table}[H]
\caption{Statistics of transition labels in the \MethodName{} Stage~2 and evaluation splits.}
\label{tab:webscene_data_statistics}
\centering
\small
\setlength{\tabcolsep}{4pt}
\resizebox{0.65\linewidth}{!}{
\begin{tabular}{lrrrr}
\toprule
Split & Samples & changed & near\_noop & exact\_noop \\
\midrule

Stage 2
& 31,245
& 25,007 
& 4,296 
& 1,942 \\

Validation
& 500
& 450 
& 45 
& 5  \\

ID test
& 1,000
& 900 
& 90 
& 10 \\

OOD test
& 1,000
& 900
& 90 
& 10 \\

\bottomrule
\end{tabular}
}
\end{table}

\begin{table}[H]
\caption{
URL and domain statistics across data splits.
}
\label{tab:domain_statistics}
\centering
\small
\setlength{\tabcolsep}{5pt}
\begin{tabular}{lccccc}
\toprule
Split
& Samples
& Unique URLs
& Max URL Frequency
& Domains
& Max Samples per Domain \\
\midrule
Stage 1
& 100,000
& 17,367
& 762
& 886
& 1,864 \\

Stage 2
& 31,245
& 4,776
& 953
& 727
& 1,771 \\

Validation
& 500
& 500
& 1
& 441
& 5 \\

ID Test
& 1,000
& 1,000
& 1
& 441
& 10 \\

OOD Test
& 1,000
& 1,000
& 1
& 303
& 31 \\
\bottomrule
\end{tabular}
\end{table}

\begin{table}[H]
\caption{
Action distribution across data splits. Numbers indicate percentages (\%).
}
\label{tab:action_distribution}
\centering
\small
\setlength{\tabcolsep}{5pt}
\begin{tabular}{lrrrrr}
\toprule
Split & click & text\_key & scroll & navigation & other \\
\midrule
Stage 2
& 35.40 & 29.64 & 23.79 & 8.13 & 3.04 \\

Validation
& 20.00 & 20.00 & 20.00 & 20.00 & 20.00 \\

ID Test
& 20.00 & 20.00 & 20.00 & 20.00 & 20.00 \\

OOD Test
& 20.00 & 20.00 & 20.00 & 20.00 & 20.00 \\
\bottomrule
\end{tabular}
\end{table}

We filter samples with missing screenshots, unparseable actions, invalid coordinate parameters, invalid TSM states, or more than 5,000 Unicode characters. Because Stage~1 has single screenshots for TSM description rather than action transitions, We report transition-type statistics only for Stage~2 and the evaluation splits.
Transitions are \texttt{exact\_noop} when screenshot hashes match; otherwise
they are \texttt{near\_noop} if grayscale SSIM $\ge 0.98$, RGB-MAE $\le 0.002$,
and Edge-F1 $\ge 0.98$ (thresholds chosen by manual inspection of minor
changes), and \texttt{changed} otherwise.
We retain no-op transitions to avoid the prior that every action changes the
interface (Table~\ref{tab:webscene_data_statistics}).

The \texttt{url} field stores the original URL, while \texttt{canonical\_domain} is obtained by lowercasing the hostname and is used for domain isolation and statistics. Table~\ref{tab:domain_statistics} summarizes the distributions of original URLs and canonical domains. \textit{Max URL frequency} denotes the maximum number of occurrences of an identical URL within a split, and \textit{Max samples per domain} denotes the maximum number of samples from a single canonical domain. The canonical domains in the OOD test set are disjoint from those in all other splits.

We group actions into five types: click, text\_key, scroll, navigation, and other (Table~\ref{tab:action_distribution}). Stage~2 follows the natural mix of the source trajectories, while the validation and test splits are stratified to 20\% per type. Table~\ref{tab:subaction_distribution} breaks each type into sub-actions on the evaluation splits.

\begin{table}[t]
\caption{
Sub-action distribution in the evaluation splits. Numbers indicate percentages (\%).
}
\label{tab:subaction_distribution}
\centering
\small
\setlength{\tabcolsep}{5pt}
\begin{tabular}{llrrr}
\toprule
Action Type & Sub-Action & Validation & In-Domain & Out-of-Domain \\
\midrule

click
& left\_click
& 15.00
& 15.00
& 15.90 \\

&
double\_click
& 2.60
& 2.50
& 1.20 \\

&
right\_click
& 2.40
& 2.50
& 2.90 \\

\midrule
text\_key
& type\_text
& 6.60
& 6.70
& 9.20 \\

&
key\_press
& 6.80
& 6.60
& 9.10 \\

&
hotkey
& 6.60
& 6.70
& 1.70 \\

\midrule
scroll
& scroll
& 20.00
& 20.00
& 20.00 \\

\midrule
navigation
& navigate
& 18.60
& 18.70
& 18.30 \\

&
page\_focus
& 1.40
& 1.30
& 1.70 \\

\midrule
other
& drag
& 10.00
& 10.00
& 14.60 \\

&
hover
& 10.00
& 10.00
& 5.40 \\

\bottomrule
\end{tabular}
\end{table}

\subsection{Focus Annotation}
\label{app:focus-annotation}

Focus annotations exist only for \texttt{changed} transitions. Given states $S_t$ and $S_{t+1}$, we mark \texttt{add} and \texttt{update} elements in $S_{t+1}$ and record their spans to compute the Base--Focus Element Loss.
Directly comparing $S_t$ and $S_{t+1}$ may confuse rendering noise or minor positional shifts with actual changes. We therefore match elements between the two TSM states. For each element, we parse content, type, state, and bbox, and record its span in the TSM string.
For text-input actions, we first match the element at the action location. If multiple nearby elements share a similar shift, we keep them matched. We then match elements with aligned semantics and bboxes, then those with aligned semantics but shifted bboxes. Unmatched elements are labeled \texttt{add} only if they pass a reliability filter. A matched element is \texttt{update} if any of content, state, or bbox changes reliably.

\section{Online Mind2Web Tasks}
\label{app:stable109}

Tables~\ref{tab:om2w-easy}--\ref{tab:om2w-hard} list the 109 Online-Mind2Web tasks used in our experiments, with official Task IDs and instructions.

\begingroup
\small
\setlength{\tabcolsep}{3pt}
\setlength{\LTleft}{\fill}
\setlength{\LTright}{\fill}
\begin{longtable}{@{}>{\centering\arraybackslash}p{0.7cm}>{\ttfamily\scriptsize\raggedright\arraybackslash}p{2.6cm}>{\raggedright\arraybackslash}p{8.5cm}@{}}
\caption{Online-Mind2Web Easy tasks.}
\label{tab:om2w-easy}\\
\toprule
\# & Task ID & Instruction \\
\midrule
\endfirsthead
\caption[]{Online-Mind2Web Easy tasks (continued).}\\
\toprule
\# & Task ID & Instruction \\
\midrule
\endhead
\midrule
\multicolumn{3}{r}{\textit{Continued on next page}} \\
\endfoot
\bottomrule
\endlastfoot
1 & 0632e496\allowbreak{}d37badee\allowbreak{}0350dad3\allowbreak{}58f047c5 & Browse recipes for gluten-free chocolate chip cookies that can be made without nuts. \\ \midrule
2 & 16200f51\allowbreak{}d63f0a47\allowbreak{}a58fa17a\allowbreak{}cd49e368 & Find a recipe that includes eggplant and mushrooms. \\ \midrule
3 & bb314cb8\allowbreak{}0f0f8489\allowbreak{}135cbf59\allowbreak{}074d11e2 & Open the page for the first Best Paper Award video recording of talks from ICLR 2016. \\ \midrule
4 & 34ccd15a\allowbreak{}8ea8fd38\allowbreak{}95af83f5\allowbreak{}ccf62369 & Find out what to do when I lose an item on a bus on us.megabus. \\ \midrule
5 & f00e7acc\allowbreak{}fb4a5e09\allowbreak{}680bdb32\allowbreak{}6e6274ad & Check the hourly forecast for Boston. \\ \midrule
6 & c521933d\allowbreak{}ad9c0ef9\allowbreak{}f1dfa2f3\allowbreak{}8b8e4405 & See the monthly forecast for Atlanta, GA. \\ \midrule
7 & 26810ed9\allowbreak{}c123a629\allowbreak{}92e3eed3\allowbreak{}1db3c5ee & Show daily weather for New York City. \\ \midrule
8 & 4e0f5561\allowbreak{}a76478da\allowbreak{}87995dee\allowbreak{}00b09572 & Show me the monthly weather forecast for Florida City. \\ \midrule
9 & d71be72a\allowbreak{}a25c3eab\allowbreak{}8eea47a0\allowbreak{}e60382e2 & Find technical specs for the latest Macbook Air on Apple. \\ \midrule
10 & f2097f92\allowbreak{}a10d42a8\allowbreak{}42c14179\allowbreak{}f422311e & Add a \$50 Uber gift card to the cart. \\ \midrule
11 & a11ecdff\allowbreak{}735b5137\allowbreak{}2d536c86\allowbreak{}6011af6f & Explore courses related to Psychology. \\ \midrule
12 & 9829f308\allowbreak{}7ab1f9c8\allowbreak{}eba6b6dd\allowbreak{}2b831d25\allowbreak{}\_112325 & Can you play the recap from the most recent NBA game? \\ \midrule
13 & 783ce6a3\allowbreak{}499fa7cf\allowbreak{}25bc12f8\allowbreak{}f0ecbbbb & Find Florida internship programs in the Mayo Clinic College of Medicine and Science. \\ \midrule
14 & 96afb3c5\allowbreak{}1146b0c2\allowbreak{}a9c55f03\allowbreak{}9a5ea6d6 & Find the most frequent word that rhymes with "thought" and has three syllables. \\ \midrule
15 & 3f312ae3\allowbreak{}efc3c3e9\allowbreak{}0ababe05\allowbreak{}0dd4e7ae & Find the current NFL standings for the AFC East division on NFL.com and go to the page on which team is in first place. \\ \midrule
16 & 3621b099\allowbreak{}326c7aeb\allowbreak{}d2e2dac6\allowbreak{}be3b52d1 & Open the profile page of the leader of the Nvidia Learning and Perception Lab. \\ \midrule
17 & 461ab9b0\allowbreak{}c7b20ac5\allowbreak{}f9127044\allowbreak{}80979c65 & Find the NYSE Rule 605 Market Center Files data for July 2024. \\ \midrule
18 & 52efbab5\allowbreak{}20734ef9\allowbreak{}bf7c09ba\allowbreak{}0f62cdc8 & Find the app for iOS. \\ \midrule
19 & c09721cc\allowbreak{}937d4dcf\allowbreak{}b391a0bc\allowbreak{}2c574b28 & Find the next available date for Albion Basin. \\ \midrule
20 & e3ab665e\allowbreak{}01e7632c\allowbreak{}e33ac1ae\allowbreak{}ca14aff6 & Find the next available dates for Alley Creek Camp. \\ \midrule
21 & 246d654f\allowbreak{}ab7c31d9\allowbreak{}651007e3\allowbreak{}9e75f74f & Open the most helpful 5-star reviews of Alpine Ridge. \\ \midrule
22 & 73d08420\allowbreak{}706ae205\allowbreak{}a9c5be28\allowbreak{}b6d4e80f & Show me the rules and cancellation for Alley Spring. \\ \midrule
23 & 95cad96f\allowbreak{}2e43f3c0\allowbreak{}d8efad13\allowbreak{}31c77c8c & View the list of the Most Popular TV on rotten tomatoes. \\ \midrule
24 & 20a460a8\allowbreak{}fe1971b8\allowbreak{}4411c5b1\allowbreak{}e6ac4186 & Show theatre events for Las Vegas and select one. \\ \midrule
25 & 92160852\allowbreak{}a6bbbc16\allowbreak{}5cee4e14\allowbreak{}ab0b1d59 & Find the shipping cost of a Common medium-sized box in flat-rate shipping and compare it with other parcel services. \\
\end{longtable}
\endgroup

\begingroup
\small
\setlength{\tabcolsep}{3pt}
\setlength{\LTleft}{\fill}
\setlength{\LTright}{\fill}
\begin{longtable}{@{}>{\centering\arraybackslash}p{0.7cm}>{\ttfamily\scriptsize\raggedright\arraybackslash}p{2.6cm}>{\raggedright\arraybackslash}p{8.5cm}@{}}
\caption{Online-Mind2Web Medium tasks.}
\label{tab:om2w-medium}\\
\toprule
\# & Task ID & Instruction \\
\midrule
\endfirsthead
\caption[]{Online-Mind2Web Medium tasks (continued).}\\
\toprule
\# & Task ID & Instruction \\
\midrule
\endhead
\midrule
\multicolumn{3}{r}{\textit{Continued on next page}} \\
\endfoot
\bottomrule
\endlastfoot
1 & 0059adc6\allowbreak{}b12a3822\allowbreak{}305deb68\allowbreak{}929b2de8 & Find support services jobs in Bentonville, in the state of Arkansas. \\ \midrule
2 & a69d2934\allowbreak{}fe54fef1\allowbreak{}65490a5a\allowbreak{}2d95bf38 & Show me recipes for pancakes with wheat and without beetroot. \\ \midrule
3 & 5d542a7e\allowbreak{}c1fa142b\allowbreak{}a73cc87d\allowbreak{}970caf39 & Find the most cited publication at the 2022 CVPR main conference. \\ \midrule
4 & b9225088\allowbreak{}86ded315\allowbreak{}c9835457\allowbreak{}a6eb43ea & Browse tenured/tenure-track faculty positions in Computer Sciences \& Technology in California. \\ \midrule
5 & 15be0597\allowbreak{}3fba714e\allowbreak{}490cd9c8\allowbreak{}84e4f072 & Find the procedure to get the license for Athletic Trainer. \\ \midrule
6 & d1807551\allowbreak{}297ac60e\allowbreak{}caaabbd2\allowbreak{}a2ed301a\allowbreak{}\_110325 & Show me a Dallas-based divorce lawyer who was once included in Top 50 Women Texas Super Lawyers list. \\ \midrule
7 & 07bdc595\allowbreak{}306729a0\allowbreak{}28ba06cc\allowbreak{}7451a80a & Select a high speed train ticket with a departure time before 23:00  from Shanghai to Beijing. \\ \midrule
8 & 63d6866f\allowbreak{}c000fcb1\allowbreak{}f153e076\allowbreak{}04bd1395 & What are the Nearby Attractions from the most popular attraction in Hong Kong? \\ \midrule
9 & 644a856c\allowbreak{}3897665e\allowbreak{}475e0dce\allowbreak{}50bf217d & Find a pair of wireless headphones on Amazon with active noise canceling for \$100 or less and add them to the cart. \\ \midrule
10 & 7072d094\allowbreak{}36972a5d\allowbreak{}5fe7476e\allowbreak{}3e9f1559 & Show me the comparison of the first two personal credit cards that do not charge foreign transaction fees. \\ \midrule
11 & 8689af4d\allowbreak{}33ce00bf\allowbreak{}2cdd8987\allowbreak{}d3bbfd86 & Add the cheapest certified refurbished iPad Air with 256GB of storage in any shade of blue to my bag. \\ \midrule
12 & ab6ee3b8\allowbreak{}3aab6cd2\allowbreak{}83320f5e\allowbreak{}01003cff & Find the tech specs of the MacBook Pro 16-inch introduced in November 2023. \\ \midrule
13 & 4d3157aa\allowbreak{}b34b54e5\allowbreak{}f0c4b965\allowbreak{}dfe930f3 & Show me community posts about pregnancy fever from the past 30 days. \\ \midrule
14 & d392e154\allowbreak{}c1c6ffbb\allowbreak{}26e2331c\allowbreak{}3afafc67 & Add a \$100 Best Buy gift card for a birthday to my cart. \\ \midrule
15 & 515f2e58\allowbreak{}11cfdd5e\allowbreak{}0e669e40\allowbreak{}f17886d8 & Search for a new internal M2 Samsung SSD drive between \$25 and \$200. \\ \midrule
16 & 87f4c512\allowbreak{}8e36cdb9\allowbreak{}366a138a\allowbreak{}7b61bb00 & View the speakers that are bluetooth and wireless and filter the results to only show models that are on sale and cost less than \$50. \\ \midrule
17 & 0a0fa834\allowbreak{}ce41b529\allowbreak{}7c647429\allowbreak{}3383759d\allowbreak{}\_110325 & What are the onboard activities of the cheapest cruise, departing from Los Angeles that lasts at least 8 days? \\ \midrule
18 & 01abae96\allowbreak{}08f2d875\allowbreak{}2a83e08f\allowbreak{}136f720c & Show me the code for the company that is the top mover in the Cboe Europe Technology Sector Index (BEPTEC) as of the latest market close. \\ \midrule
19 & a48e2f1e\allowbreak{}e8d87eae\allowbreak{}ea56fe5e\allowbreak{}730427e6 & Pass the first trending chess puzzle. \\ \midrule
20 & ba2a469a\allowbreak{}f584f16d\allowbreak{}a93ce6a7\allowbreak{}430cf7e5 & Search for a beginner’s course in computer science that includes advertisement skills. \\ \midrule
21 & c698ff3f\allowbreak{}c0f6cbce\allowbreak{}39947c59\allowbreak{}7ab5749b\allowbreak{}\_121425 & Help me find a list of school activity events in New York for this month. \\ \midrule
22 & 608c595e\allowbreak{}ec271fa5\allowbreak{}dc035069\allowbreak{}23519994 & Calculate a FedEx Ground shipping rate for a 3-pound package from zip code 10019 to zip code 90028. \\ \midrule
23 & a8b9edd5\allowbreak{}98561d2d\allowbreak{}e901864d\allowbreak{}5f40fe67 & Calculate the shipping cost for 4 pound package from Texas to New York. \\ \midrule
24 & 6ca20f1d\allowbreak{}a01edeb4\allowbreak{}9a7a42c8\allowbreak{}16d8c6fe & Find the Eligibility to get the child benefit and How it works and how to claim \\ \midrule
25 & b320c68b\allowbreak{}ffc1f3c7\allowbreak{}f2a8dc9d\allowbreak{}5478fb27 & Find a walkthrough for the game "The Legend of Zelda: Breath of the Wild" on ign. \\ \midrule
26 & aa4b5cb7\allowbreak{}114fcc13\allowbreak{}8ade82b4\allowbreak{}b9716d24 & Find an editor's choice review with a score of 10 in the boardgame category on ign. \\ \midrule
27 & da8f3823\allowbreak{}a827c7d3\allowbreak{}a492f383\allowbreak{}808e7912 & Find and open the earliest press release. \\ \midrule
28 & 2c20d87a\allowbreak{}046fadcb\allowbreak{}6ff07ee8\allowbreak{}77bfbf37 & Open the form 8843 for tax year 2022. \\ \midrule
29 & 65c4030f\allowbreak{}22fb6eb1\allowbreak{}01acfee4\allowbreak{}825f1318 & Find a female MD Cardiologist in Jacksonville, Florida. \\ \midrule
30 & c94551d2\allowbreak{}b18f9ad0\allowbreak{}ab31b0bd\allowbreak{}98ca42e3\allowbreak{}\_121125 & Find cats available for adoption within 25 miles of zip code 94587, Young or adult-age cats, sorted by Oldest Addition. \\ \midrule
31 & b1ce968a\allowbreak{}361e1088\allowbreak{}ce8d2ade\allowbreak{}6c2c9af0 & Find young cats in Seattle and show off the newest additions. \\ \midrule
32 & 2fc51dd3\allowbreak{}febd447f\allowbreak{}0fdcdabc\allowbreak{}a8d944ce\allowbreak{}\_110325 & Locate a self-storage unit near zip code 60538 that can fit the contens of two small bedrooms and is climate-controlled. \\ \midrule
33 & 502e8644\allowbreak{}40283214\allowbreak{}e0180645\allowbreak{}015f568b\allowbreak{}\_110325 & Check permit availability for a group of 4 in Brooks Camp, Katmai National Park for next Saturday. \\ \midrule
34 & 6174e5dd\allowbreak{}d40cfbdc\allowbreak{}33ee1502\allowbreak{}f40bac39 & Find a day-use park that offers horseback riding near Nashville. \\ \midrule
35 & 47e314cc\allowbreak{}452c5405\allowbreak{}24ffb7cf\allowbreak{}520285a3 & Find the park that offers the cheapest paddling permits. \\ \midrule
36 & 78f39733\allowbreak{}6b6fd1cb\allowbreak{}ba0127db\allowbreak{}7a8cd502 & Browse the upcoming SuperBike events taking place in Italy. \\ \midrule
37 & 753f372c\allowbreak{}189d3b30\allowbreak{}6623cb0c\allowbreak{}65b50320 & Compare the U.S. ETP Odd Lot Rate (\%) between Quartile 1 and Quartile 4, viewing quartiles by price, and display the chart with a logarithmic scale on the vertical axis. \\ \midrule
38 & 9d09bc94\allowbreak{}8462db03\allowbreak{}2bac9896\allowbreak{}8b11b008 & Find NHL events occurring in Boston. \\ \midrule
39 & 85b284c1\allowbreak{}8d7e78c9\allowbreak{}b5a9e074\allowbreak{}e7aa3b98\allowbreak{}\_051526 & View the cheapest apartment available for students at New York University with bills that include WIFI and cleaning services. \\ \midrule
40 & 733f1d8b\allowbreak{}f79d5bc2\allowbreak{}240c5357\allowbreak{}f928ffff & Find the cheapest travel deal or discount to Thailand that lasts more than 10 days, departs in next month, and show the total price. \\ \midrule
41 & 9b5dfe54\allowbreak{}a1c14c5c\allowbreak{}6336bae7\allowbreak{}374c3bb5 & Find a UPS Access Point near SPRING, TX and services provided by them. \\ \midrule
42 & 9c04b71b\allowbreak{}b8db6cf8\allowbreak{}e743b229\allowbreak{}0cbc8797 & Find a UPS drop-off point near Miami Florida. \\ \midrule
43 & 9d46ccb9\allowbreak{}15eff39e\allowbreak{}e1ae1e73\allowbreak{}28f5f20d & Get a quote for the fastest shipping available for 5 lbs with dimensions of 4 inches in length, width, and height from New York, NY 10001, USA to Truckee, California 96162, USA. \\ \midrule
44 & 3084bc22\allowbreak{}5219fcb7\allowbreak{}3dc1cb0f\allowbreak{}97276c1c & Get quotes for a package weighing 10 lbs with dimensions of 2 inches in length, width, and height, being shipped from Long Beach, 90802 to Portland, 97201. \\ \midrule
45 & 7abdceee\allowbreak{}212151f1\allowbreak{}87ee1a17\allowbreak{}44c57606 & Can you show me the page with the filing fee for a self-petitioned I-140 application? \\ \midrule
46 & ce616721\allowbreak{}ce9aeda6\allowbreak{}9890fbcc\allowbreak{}b29677a6 & Calculate the price to ship a large flat-rate box from 77449 to 77084 at the first available date and time. \\ \midrule
47 & 7562d9b4\allowbreak{}e4829a44\allowbreak{}245aafce\allowbreak{}2e1f62db & Find the nearest location to zip code 54620 that offers size 4 P.O. Boxes. \\ \midrule
48 & c7c07ec1\allowbreak{}0c668625\allowbreak{}a21ba641\allowbreak{}65d719bb & Find the total monthly price for four prepaid unlimited lines without autopay discounts. \\
\end{longtable}
\endgroup

\begingroup
\small
\setlength{\tabcolsep}{3pt}
\setlength{\LTleft}{\fill}
\setlength{\LTright}{\fill}
\begin{longtable}{@{}>{\centering\arraybackslash}p{0.7cm}>{\ttfamily\scriptsize\raggedright\arraybackslash}p{2.6cm}>{\raggedright\arraybackslash}p{8.5cm}@{}}
\caption{Online-Mind2Web Hard tasks.}
\label{tab:om2w-hard}\\
\toprule
\# & Task ID & Instruction \\
\midrule
\endfirsthead
\caption[]{Online-Mind2Web Hard tasks (continued).}\\
\toprule
\# & Task ID & Instruction \\
\midrule
\endhead
\midrule
\multicolumn{3}{r}{\textit{Continued on next page}} \\
\endfoot
\bottomrule
\endlastfoot
1 & f158345f\allowbreak{}8489e0d1\allowbreak{}d91e2876\allowbreak{}8c39bca1 & Estimate the total cost (with basic support) of using 5 million input tokens and 5 million output tokens each for GPT-4o and GPT-4o Mini, both deployed in the US/EU Data Zones under Standard (On-Demand) in the East US region. \\ \midrule
2 & d1970c16\allowbreak{}271496cb\allowbreak{}be166ecb\allowbreak{}ecc0a1d8 & I'm 25 and located in Texas. Shop for 2020 made dry red wine made in United States priced between 15-20 dollars and add 5 bottles to the cart. \\ \midrule
3 & 1ab384fb\allowbreak{}3a791edf\allowbreak{}b410213c\allowbreak{}c6b82151\allowbreak{}\_110325 & Show me the result of a proton emission decay for a Beryllium nucleus with 4 protons and 2 neutrons in the simulation. \\ \midrule
4 & 824eb7bb\allowbreak{}0ef1ce40\allowbreak{}bfd49c12\allowbreak{}182d9428\allowbreak{}\_110325 & Get the women's black one-piece swimsuit, size large, with the highest discount percentage, and add it to my cart. \\ \midrule
5 & a5c87cc1\allowbreak{}c94a090c\allowbreak{}9a8dc2c8\allowbreak{}b6a125d0 & Find the SO2 air quality over the past hour for Maine North, County Cork, Ireland. \\ \midrule
6 & a13e4231\allowbreak{}a3d6a700\allowbreak{}0c622c56\allowbreak{}448d97ba & Find an Airbnb in Cleveland for three nights. The check-in date is the day after tomorrow. We have 2 adults, 2 kids, and 1 pet. The budget is \$100 to \$300 per night. Essential amenities include free parking, a washer, and a gym. \\ \midrule
7 & 1b867afe\allowbreak{}cf072cb8\allowbreak{}77ebfa40\allowbreak{}69263746 & Display the figure comparing unemployment trends among women in Illinois and Michigan. \\ \midrule
8 & d4fb78b7\allowbreak{}e74508cd\allowbreak{}3b33f01c\allowbreak{}f9200997 & Show the figure comparing Occupational Fatalities Trends between Ohio and New York. \\ \midrule
9 & 50d91eab\allowbreak{}de542906\allowbreak{}937ab4c5\allowbreak{}b6f8f23a\allowbreak{}\_121425 & My baby boy was born on Dec 1, 2025. His weight is 7 lb 8 oz, height is 20 inches, and head circumference is 35 cm. The measurement date is Dec 1, 2025. Can you calculate his growth percentiles using a calculator? \\ \midrule
10 & 7680a920\allowbreak{}359cb1a5\allowbreak{}08fbddb0\allowbreak{}01b98167 & See the prediction about the girl child's height, whose current height at seven years is 4 feet and whose weight is 55 lbs, her mother is 5 feet 2, and her father is 5 feet 8. \\ \midrule
11 & 4464a842\allowbreak{}1f8bc878\allowbreak{}6524a499\allowbreak{}258dfad3 & Check the specifications of the best-selling HP FHD laptop with 16 GB RAM and core i7 running on Windows 11. \\ \midrule
12 & 7e1047f4\allowbreak{}803237f3\allowbreak{}19c004f7\allowbreak{}a7f6bccb\allowbreak{}\_121125 & Discover the trade-in value of my Intel 10th generation i5, 2021, HP laptop in fair condition, which has 8 GB memory and can be powered on, proceed for the in-store trade-in. \\ \midrule
13 & fc53ddd3\allowbreak{}421411a4\allowbreak{}1c1020a3\allowbreak{}fdc84ec4\allowbreak{}\_051526 & I want to purchase an open-box Samsung Galaxy S25 Plus in excellent condition and trade in a gray Galaxy S20 5G (Verizon), with a perfect screen, in good condition. How much would it cost? \\ \midrule
14 & 5dec0e66\allowbreak{}20849459\allowbreak{}f29e6465\allowbreak{}982c597e & Search for 33 to 49inch Qled gaming monitor with a 240hz refresh rate that is within \$1000 to \$2000. \\ \midrule
15 & b2f4fde2\allowbreak{}fce122a9\allowbreak{}3c7b5780\allowbreak{}86cb0585 & Find the cheapest hotel + flight + car package from New York to San Francisco, departing tomorrow and returning on the fourth day from departure, for two adults and a six-year-old child. The package should be one room with free breakfast and spa access. \\ \midrule
16 & 0e5536aa\allowbreak{}ad9d3462\allowbreak{}b06cf725\allowbreak{}e6ed535a & Show me the page with average wait times for U.S. citizens arriving at Raleigh-Durham International Airport on 2025-03-12. \\ \midrule
17 & 9ed38272\allowbreak{}66b3b804\allowbreak{}f485859c\allowbreak{}3d00401e & If I'm 30, plan to retire at 65, and can save \$300/month, with a 3\% annual return, 13\% current tax rate, and 24\% retirement tax rate, show the comparison chart between Traditional and Roth IRA. \\ \midrule
18 & eb323dc5\allowbreak{}84156d0e\allowbreak{}b3a2b90b\allowbreak{}b8c4b791\allowbreak{}\_110325 & Find the cheapest 2 bed and 3+ bath apartment listing for rent in New York. \\ \midrule
19 & 1223b075\allowbreak{}36a87e01\allowbreak{}70ff87cb\allowbreak{}bebd1d3c & Complete a multiplication quiz on https://www.coolmath4kids.com/, covering multiplication facts for 11-12. The quiz should consist of 10 questions, with unlimited time allowed for each. The goal is to achieve a perfect score of 10 out of 10. \\ \midrule
20 & 949dc965\allowbreak{}a6c23a95\allowbreak{}663b3bc2\allowbreak{}ca2c3a8a & Find UA or AA flights from London to New York that arrive between 8:00 PM and 11:00 PM on FlightAware. \\ \midrule
21 & 0170ca95\allowbreak{}038b05fa\allowbreak{}58d463fe\allowbreak{}627ac605 & Check if a visa is required to work in the UK for longer than 6 months in Healthcare as an American citizen. \\ \midrule
22 & ec78d3a6\allowbreak{}35e417bc\allowbreak{}2a80d03c\allowbreak{}a93d7165 & What are the benefits and financial support a single person living in England, over the state pension age, unemployed, with no health conditions, or caring for someone with one, can get? \\ \midrule
23 & c6c9dc60\allowbreak{}79677cef\allowbreak{}594cec2f\allowbreak{}a6b16602 & Add the cheapest black sofa with at least three seats, a leather finish, and at least four stars to my cart. \\ \midrule
24 & 1c3b747a\allowbreak{}e12ccee8\allowbreak{}95745f82\allowbreak{}e3f2ef8a & Identify the ongoing competition that offers the highest prize and find the code that received the most votes in that competition. \\ \midrule
25 & b4aa7315\allowbreak{}e31dfcdc\allowbreak{}52baf777\allowbreak{}1be260c9 & Find the HGX H100 driver for Ubuntu 22.04 on AMD64 CPU. \\ \midrule
26 & 0b2623e9\allowbreak{}fa5cea99\allowbreak{}7f76490b\allowbreak{}cbc5220f & Find a list of shorthaired dogs available for adoption within 100 miles of zip code 94587 that are good with kids and cats, and have been on Petfinder for over 30 days. \\ \midrule
27 & f707d765\allowbreak{}bca66883\allowbreak{}0745d208\allowbreak{}07d7bee6 & Show me the list of young female English Spot rabbits available for adoption in Chicago, IL, within 50 miles. \\ \midrule
28 & c3a33396\allowbreak{}8fc3c43d\allowbreak{}7f2688f4\allowbreak{}25a0d633 & Find the cheapest certified pre-owned Porsche 911 with a model year of 2019 or newer, within a 200-mile radius of ZIP code 97007. \\ \midrule
29 & 59912927\allowbreak{}c1fddee6\allowbreak{}ded8a499\allowbreak{}86896bc2 & Look for the most useful reviews of the highest-rated anti-reflective TVs with screen sizes from 55" to 64" and prices ranging from \$300 to \$1500. \\ \midrule
30 & 27fa3ac2\allowbreak{}0745d3d3\allowbreak{}5e89fae1\allowbreak{}57f63069 & Browse the class schedule of graduate-level chemistry courses on Monday afternoons in the winter of 2023. \\ \midrule
31 & 78baf9db\allowbreak{}e7c3532f\allowbreak{}7d7ef4cc\allowbreak{}22a7f065\allowbreak{}\_051526 & Find the latest available digital trends to market insights in China (2026 focus). \\ \midrule
32 & 1bc15437\allowbreak{}7120ec15\allowbreak{}b18dbabd\allowbreak{}ba49c741\allowbreak{}\_051526 & Book 4 tickets in the upper for any BTS show near New York in the next six months and view ticket prices with estimated fees. \\ \midrule
33 & 9d090a15\allowbreak{}c214eb07\allowbreak{}0d9caa8a\allowbreak{}034d03c1 & Find the lowest-priced Student housing near Liverpool International College which has been priced between 100 to 300 pounds and has a private bathroom. \\ \midrule
34 & 8ae51035\allowbreak{}5d978424\allowbreak{}f490798f\allowbreak{}900bfa2c\allowbreak{}\_051526 & Show me the private rooms in New York that has a private bathroom, wifi, and gas included in the bills. \\ \midrule
35 & bb5d90e6\allowbreak{}f2fbc0ae\allowbreak{}146f7c19\allowbreak{}98c2b4a1 & Find the most viewed TED talk on the topic of robots that lasts between 12 and 18 minutes. \\ \midrule
36 & b7a9a6b5\allowbreak{}d451164c\allowbreak{}09bbd27b\allowbreak{}670bc2ae & Show me the list of Men's Blazers, Black, Size M on Uniqlo. \\
\end{longtable}
\endgroup

\section{Prompts}
\label{app:prompts}

\subsection{Web Agent Generates Candidate Actions}
\begin{tcolorbox}[
  breakable,
  colback=gray!6,
  colframe=black!25,
  boxrule=0.4pt,
  arc=1pt,
  left=3pt,right=3pt,top=2pt,bottom=2pt
]
\begin{lstlisting}[style=prompt]
You are a GUI Agent operating a web browser from the current screenshot. Generate candidate actions that advance the task.

## Input

- `goal_text`: final user task.
- `current_url`: current browser URL.
- `current_subgoal_trace`: actions already attempted.
- `supported_unified_actions`: allowed action types.
- The attached image is the current viewport screenshot.

## Candidate policy

- Return one direct-execution candidate at confidence 8-10.
- Return two or three complementary candidates at confidence 0-7, including exploratory actions for unfamiliar interfaces.
- Set `confidence` to an integer from 0 to 10 indicating confidence that the candidate is the correct next action.

## Action contract

- Express coordinates on a `0..999` grid relative to the screenshot.
- Choose action types from `supported_unified_actions`.
- Ground coordinate actions in visible screenshot evidence.
- Use `type_text` for a focused input; use a click candidate to establish focus.
- Use `finish` when the screenshot or action history shows task completion.

Use the parameters defined for each action type:

- `left_click`, `double_click`, `right_click`, `toggle_check`, `hover`, `move_to`: `{"type":"...","coordinate":[x,y]}`.
- `type_text`: `{"type":"type_text","text":"..."}`; `coordinate`, `clear`, and `enter` are optional.
- `scroll`: `{"type":"scroll","delta_x":0,"delta_y":500}` with integer deltas in `[-999,999]`; positive `delta_y` scrolls down.
- `key_press`: `{"type":"key_press","key":"Enter"}`.
- `hotkey`: `{"type":"hotkey","keys":["ctrl","l"]}`.
- `navigate`: `{"type":"navigate","url":"https://..."}` with an absolute HTTP(S) URL.
- `go_back`, `go_forward`, `refresh`, `wait`: `{"type":"..."}`.
- `select_option`: `{"type":"select_option","coordinate":[x,y],"option_text":"..."}`.
- `drag`: `{"type":"drag","from_coordinate":[x1,y1],"to_coordinate":[x2,y2]}`.
- `clear`: `{"type":"clear"}` or `{"type":"clear","coordinate":[x,y]}`.
- `finish`: `{"type":"finish","answer":"..."}`.
- `fail`: `{"type":"fail","reason":"..."}`.

## Output

Return a valid JSON object in this format:

{
  "candidate_actions": [
    {
      "unified_action": {
        "type": "left_click",
        "coordinate": [500, 500]
      },
      "action_intent": "Open the filter panel",
      "confidence": 8
    }
  ]
}

\end{lstlisting}
\end{tcolorbox}

\subsection{Web Agent Digest}
\begin{tcolorbox}[
  breakable,
  colback=gray!6,
  colframe=black!25,
  boxrule=0.4pt,
  arc=1pt,
  left=3pt,right=3pt,top=2pt,bottom=2pt
]
\begin{lstlisting}[style=prompt]
You are a GUI Agent in Digest mode. Analyze one candidate action and convert its World Model forecast into a concise, grounded record for the action selector.

## Input

- `goal_text`: final user task.
- The attached image is the current viewport screenshot.
- `candidate_action`: one proposed unified action.
- `candidate_action_intent`: the candidate's intended visible effect.
- `predicted_next_md`: the complete predicted Markdown state after executing this candidate.

## Action contract

Coordinates in candidate_action use the 0..999 grid.
Use the parameters defined for each action type:

- `left_click`, `double_click`, `right_click`, `toggle_check`, `hover`, `move_to`: `{"type":"...","coordinate":[x,y]}`.
- `type_text`: `{"type":"type_text","text":"..."}`; `coordinate`, `clear`, and `enter` are optional.
- `scroll`: `{"type":"scroll","delta_x":0,"delta_y":500}` with integer deltas in `[-999,999]`; positive `delta_y` scrolls down.
- `key_press`: `{"type":"key_press","key":"Enter"}`.
- `hotkey`: `{"type":"hotkey","keys":["ctrl","l"]}`.
- `navigate`: `{"type":"navigate","url":"https://..."}` with an absolute HTTP(S) URL.
- `go_back`, `go_forward`, `refresh`, `wait`: `{"type":"..."}`.
- `select_option`: `{"type":"select_option","coordinate":[x,y],"option_text":"..."}`.
- `drag`: `{"type":"drag","from_coordinate":[x1,y1],"to_coordinate":[x2,y2]}`.
- `clear`: `{"type":"clear"}` or `{"type":"clear","coordinate":[x,y]}`.
- `finish`: `{"type":"finish","answer":"..."}`.
- `fail`: `{"type":"fail","reason":"..."}`.

## Markdown Grammar

Markdown uses the same `0..999` relative coordinate grid as `candidate_action`.
Each UI element uses center-box coordinates:

`[content](type)<cx,cy,w,h>`

Meaning:
- `cx,cy` are the element center point.
- `w,h` are the element width and height.
- All four numbers are integers on the `0..999` grid.

Valid line formats:
- `[text](text)<cx,cy,w,h>`
- `[text](link)<cx,cy,w,h>`
- `[text](button)<cx,cy,w,h>`
- `[visible value or placeholder](input)<cx,cy,w,h>`
- `[icon: semantic description](icon)<cx,cy,w,h>`
- `[img: semantic description](image)<cx,cy,w,h>`
- `[ ] [label](toggle)<cx,cy,w,h>`
- `[x] [label](toggle)<cx,cy,w,h>`
- `[>] [label](select)<cx,cy,w,h>`
- `[v] [label](select)<cx,cy,w,h>`
- `- [Block: description]<cx,cy,w,h>`
- `[Environment: Web]`

Element type definitions:
- `text`: visible non-interactive text.
- `link`: visible text or content that navigates when activated.
- `button`: An element with a visible border and a color that clearly contrasts with the webpage background. Its content is usually text, and it is generally larger than an icon.
- `input`: a visible editable field; use its current value, otherwise its visible placeholder.
- `icon`: A small graphical symbol whose background color usually matches the webpage background. It may optionally have a black outline.
- `image`: A photograph or visual depicting natural scenes, people, or similar content, with boundaries clearly distinguishable from the webpage background.
- `toggle`: Any checkable element, such as a checkbox or radio button.
- `select`: A dropdown or expandable selection control.



### Toggle State

- `[ ]` indicates unchecked, off, or unselected.
- `[x]` indicates checked, on, or selected.
- Clicking a toggle usually switches between `[ ]` and `[x]`.
- When a toggle state changes, keep the same label and coordinates.

### Select State

- `[>]` indicates collapsed or closed.
- `[v]` indicates expanded or open.
- Clicking a closed select usually expands it and displays options nearby.
- Clicking an open select closes it.

### Block

Block description must be one of:
- `Navigation`
- `Sidebar`
- `Footer`
- `Form`
- `Table`
- `Section`

## Digest rules

1. Use the screenshot as the current state and `predicted_next_md` as the forecasted next state.
2. Compare both states and summarize the single most task-relevant visible change.
3. Evaluate `candidate_action_intent` as a hypothesis against visible forecast evidence.
4. Set `intent_alignment` to:
   - `yes`: the forecast visibly supports the intended effect;
   - `partial`: the forecast visibly supports part of the intended effect;
   - `no`: the forecast visibly contradicts the intended effect;
   - `unclear`: the forecast leaves intent alignment ambiguous.
5. Set `task_progress` to:
   - `positive`: the forecast visibly advances or completes `goal_text`;
   - `neutral`: the forecast preserves the task-relevant state;
   - `negative`: the forecast visibly moves away from or blocks the task;
   - `unclear`: the forecast leaves task progress ambiguous.
6. Cite up to three short, exact excerpts from `predicted_next_md` that support the digest.
7. Infer element removal from explicit replacement or navigation evidence.
8. Map ambiguous grounded task evidence to `"predicted_change": "unclear"`, `unclear` for both labels, and an empty `evidence_refs` list.

## Output

Return valid JSON with exactly these top-level keys:

{
  "predicted_change": "One concise sentence describing the task-relevant visible change",
  "intent_alignment": "yes",
  "task_progress": "positive",
  "evidence_refs": [
    "exact substring copied from predicted_next_md"
  ]
}

- `predicted_change` contains at most 40 words.
- `intent_alignment` uses `yes`, `partial`, `no`, or `unclear`.
- `task_progress` uses `positive`, `neutral`, `negative`, or `unclear`.
- `evidence_refs` contains zero to three short exact excerpts like `[text](text)<cx,cy,w,h>`.

\end{lstlisting}
\end{tcolorbox}

\subsection{Web Agent Selects the Best Action}
\begin{tcolorbox}[
  breakable,
  colback=gray!6,
  colframe=black!25,
  boxrule=0.4pt,
  arc=1pt,
  left=3pt,right=3pt,top=2pt,bottom=2pt
]

\begin{lstlisting}[style=prompt]
You are a GUI Agent operating a web browser from the current screenshot. Select the candidate action most likely to advance the task.

Each `candidate_digests[i]` summarizes the World Model forecast for `candidate_actions[i]`.

## Input

- `goal_text`: final user task.
- `current_url`: current browser URL.
- `current_subgoal_trace`: actions already attempted.
- `supported_unified_actions`: allowed action types.
- The attached image is the current viewport screenshot.
- `candidate_actions`: candidate actions in their original order.
- `candidate_actions_intent`: intended visible effect of each candidate action.
- `candidate_digests`: Digest outputs in candidate order.

## Selection rules

1. Align actions, intents, and digests by index.
2. Use the screenshot and action history as current evidence and each digest as forecast evidence.
3. Evaluate `intent_alignment` and `task_progress` independently.
4. Rank candidates by likely useful task progress, grounded by `predicted_change`, `intent_alignment` and `task_progress`.
5. Resolve an `unclear` digest using the current state, action history, and candidate intent.
6. Prefer candidates that break repeated ineffective attempts recorded in the action history.
7. Select exactly one existing candidate and copy its action and intent by index.

## Output

Return a valid JSON object in this format:

{
  "best_action": {
    "candidate_index": 0,
    "unified_action": {
      "type": "left_click",
      "coordinate": [500, 500]
    },
    "action_intent": "Expected visible change"
  }
}

- `candidate_index` is a zero-based integer in `[0, number_of_candidates - 1]`.
- `unified_action` equals `candidate_actions[candidate_index]`.
- `action_intent` equals `candidate_actions_intent[candidate_index]`.

\end{lstlisting}
\end{tcolorbox}

\subsection{Vanilla Web Agent}
\begin{tcolorbox}[
  breakable,
  colback=gray!6,
  colframe=black!25,
  boxrule=0.4pt,
  arc=1pt,
  left=3pt,right=3pt,top=2pt,bottom=2pt
]
\begin{lstlisting}[style=prompt]
You are a GUI Agent operating a web browser from the current screenshot. Generate exactly one next action that best advances the task.

## Input

- `goal_text`: final user task.
- `current_url`: current browser URL.
- `current_subgoal_trace`: actions already attempted.
- `supported_unified_actions`: allowed action types.
- The attached image is the current viewport screenshot.

## Coordinate and action contract

- All coordinates use a `0..999` grid relative to the screenshot.
- Use only action types listed in `supported_unified_actions`.
- Coordinate actions must target visible screenshot evidence.
- `type_text` assumes the intended input is already focused; otherwise propose a click separately.
- Use `finish` only when the screenshot or action history shows the task is complete.

Each `unified_action` must use exactly the parameters for its action type:

- `left_click`, `double_click`, `right_click`, `toggle_check`, `hover`, `move_to`: `{"type":"...","coordinate":[x,y]}`.
- `type_text`: `{"type":"type_text","text":"..."}`; `coordinate`, `clear`, and `enter` are optional.
- `scroll`: `{"type":"scroll","delta_x":0,"delta_y":500}` with integer deltas in `[-999,999]`; positive `delta_y` scrolls down.
- `key_press`: `{"type":"key_press","key":"Enter"}`.
- `hotkey`: `{"type":"hotkey","keys":["ctrl","l"]}`.
- `navigate`: `{"type":"navigate","url":"https://..."}` with an absolute HTTP(S) URL.
- `go_back`, `go_forward`, `refresh`, `wait`: only `{"type":"..."}`; never add `coordinate`.
- `select_option`: `{"type":"select_option","coordinate":[x,y],"option_text":"..."}`.
- `drag`: `{"type":"drag","from_coordinate":[x1,y1],"to_coordinate":[x2,y2]}`.
- `clear`: `{"type":"clear"}` or `{"type":"clear","coordinate":[x,y]}`.
- `finish`: `{"type":"finish","answer":"..."}`.
- `fail`: `{"type":"fail","reason":"..."}`.

Before returning, validate the selected action against this table.

## Output

Output only JSON. 

{
  "selected_action": {
    "unified_action": {
      "type": "left_click",
      "coordinate": [500, 500]
    },
    "action_intent": "expected visible change"
  }
}

Requirements:
- Return exactly one `selected_action`.

\end{lstlisting}
\end{tcolorbox}

\subsection{MKD2World Reconstructs the TSM}
\begin{tcolorbox}[
  breakable,
  colback=gray!6,
  colframe=black!25,
  boxrule=0.4pt,
  arc=1pt,
  left=3pt,right=3pt,top=2pt,bottom=2pt
]
\begin{lstlisting}[style=prompt]
Given the current screenshot, generate the complete visible interface Markdown.

## Input

- `current_screenshot`: the current webpage screenshot.

## Output Format

Output Markdown directly:

[Environment: Web]
...

Rules:
- Markdown must start with `[Environment: Web]`.



## Markdown Grammar

Markdown uses a 0..999 coordinate grid relative to current_screenshot.
Each UI element uses center-box coordinates:

`[content](type)<cx,cy,w,h>`

Meaning:
- `cx,cy` are the element center point.
- `w,h` are the element width and height.
- All four numbers are integers on the `0..999` grid.

Valid line formats:
- `[text](text)<cx,cy,w,h>`
- `[text](link)<cx,cy,w,h>`
- `[text](button)<cx,cy,w,h>`
- `[visible value or placeholder](input)<cx,cy,w,h>`
- `[icon: semantic description](icon)<cx,cy,w,h>`
- `[img: semantic description](image)<cx,cy,w,h>`
- `[ ] [label](toggle)<cx,cy,w,h>`
- `[x] [label](toggle)<cx,cy,w,h>`
- `[>] [label](select)<cx,cy,w,h>`
- `[v] [label](select)<cx,cy,w,h>`
- `- [Block: description]<cx,cy,w,h>`
- `[Environment: Web]`

Element type definitions:
- `text`: visible non-interactive text.
- `link`: visible text or content that navigates when activated.
- `button`: An element with a visible border and a color that clearly contrasts with the webpage background. Its content is usually text, and it is generally larger than an icon.
- `input`: a visible editable field; use its current value, otherwise its visible placeholder.
- `icon`: A small graphical symbol whose background color usually matches the webpage background. It may optionally have a black outline.
- `image: A photograph or visual depicting natural scenes, people, or similar content, with boundaries clearly distinguishable from the webpage background.
- `toggle`: Any checkable element, such as a checkbox or radio button.
- `select`: A dropdown or expandable selection control.



### Toggle State

- `[ ]` indicates unchecked, off, or unselected.
- `[x]` indicates checked, on, or selected.

### Select State

- `[>]` indicates collapsed or closed.
- `[v]` indicates expanded or open.

### Block

Block description must be one of:
- `Navigation`
- `Sidebar`
- `Footer`
- `Form`
- `Table`
- `Section`

### Concrete Example


[Environment: Web]

- [Block: Navigation]<500,40,960,60>
  - [Home](link)<80,40,80,30>
  - [icon: menu](icon)<950,40,30,30>
- [Block: Sidebar]<110,350,180,520>
  - [ ] [Dark](toggle)<100,150,120,30>
  - [x] [Sync](toggle)<100,200,120,30>
  - [>] [More](select)<100,250,120,30>
- [Block: Form]<590,210,720,200>
  - [Query](input)<520,190,480,35>
  - [Go](button)<850,190,100,35>
- [Block: Section]<590,410,720,160>
  - [v] [Mode](select)<340,370,160,30>
  - [sample1](text)<340,415,160,30>
  - [sample2](text)<340,460,160,30>
- [Block: Table]<590,600,720,160>
  - [Row](text)<360,580,200,25>
- [Block: Footer]<500,960,960,60>
  - [img: blue mountain lake at sunset](image)<160,960,240,30>


## Rules

The top-to-bottom order of elements in the prediction should roughly match their order in the webpage viewport.
Prioritize correct semantics and element types, then infer reasonable coordinate positions.


\end{lstlisting}
\end{tcolorbox}

\subsection{MKD2World Predicts the TSM}
\begin{tcolorbox}[
  breakable,
  colback=gray!6,
  colframe=black!25,
  boxrule=0.4pt,
  arc=1pt,
  left=3pt,right=3pt,top=2pt,bottom=2pt
]
\begin{lstlisting}[style=prompt]
Given the current screenshot and one candidate GUI action, predict the complete visible interface Markdown after that action is executed.

## Input

- `current_screenshot`: screenshot before the action is executed.
- `candidate_action`: one candidate action generated by the GUI Agent.

## Output Format

Output Markdown directly:

[Environment: Web]
...

Rules:
- Markdown must be the complete visible UI state after the action.
- Markdown must start with `[Environment: Web]`.

## Markdown Grammar

Markdown uses the same `0..999` relative coordinate grid as `candidate_action`.
Each UI element uses center-box coordinates:

`[content](type)<cx,cy,w,h>`

Meaning:
- `cx,cy` are the element center point.
- `w,h` are the element width and height.
- All four numbers are integers on the `0..999` grid.

Valid line formats:
- `[text](text)<cx,cy,w,h>`
- `[text](link)<cx,cy,w,h>`
- `[text](button)<cx,cy,w,h>`
- `[visible value or placeholder](input)<cx,cy,w,h>`
- `[icon: semantic description](icon)<cx,cy,w,h>`
- `[img: semantic description](image)<cx,cy,w,h>`
- `[ ] [label](toggle)<cx,cy,w,h>`
- `[x] [label](toggle)<cx,cy,w,h>`
- `[>] [label](select)<cx,cy,w,h>`
- `[v] [label](select)<cx,cy,w,h>`
- `- [Block: description]<cx,cy,w,h>`
- `[Environment: Web]`

Element type definitions:
- `text`: visible non-interactive text.
- `link`: visible text or content that navigates when activated.
- `button`: An element with a visible border and a color that clearly contrasts with the webpage background. Its content is usually text, and it is generally larger than an icon.
- `input`: a visible editable field; use its current value, otherwise its visible placeholder.
- `icon`: A small graphical symbol whose background color usually matches the webpage background. It may optionally have a black outline.
- `image`: A photograph or visual depicting natural scenes, people, or similar content, with boundaries clearly distinguishable from the webpage background.
- `toggle`: Any checkable element, such as a checkbox or radio button.
- `select`: A dropdown or expandable selection control.



### Toggle State

- `[ ]` indicates unchecked, off, or unselected.
- `[x]` indicates checked, on, or selected.
- Clicking a toggle usually switches between `[ ]` and `[x]`.
- When a toggle state changes, keep the same label and coordinates.

### Select State

- `[>]` indicates collapsed or closed.
- `[v]` indicates expanded or open.
- Clicking a closed select usually expands it and displays options nearby.
- Clicking an open select closes it.

### Block

Block description must be one of:
- `Navigation`
- `Sidebar`
- `Footer`
- `Form`
- `Table`
- `Section`

### Concrete Example


[Environment: Web]

- [Block: Navigation]<500,40,960,60>
  - [Home](link)<80,40,80,30>
  - [icon: menu](icon)<950,40,30,30>
- [Block: Sidebar]<110,350,180,520>
  - [ ] [Dark](toggle)<100,150,120,30>
  - [x] [Sync](toggle)<100,200,120,30>
  - [>] [More](select)<100,250,120,30>
- [Block: Form]<590,210,720,200>
  - [Query](input)<520,190,480,35>
  - [Go](button)<850,190,100,35>
- [Block: Section]<590,410,720,160>
  - [v] [Mode](select)<340,370,160,30>
  - [sample1](text)<340,415,160,30>
  - [sample2](text)<340,460,160,30>
- [Block: Table]<590,600,720,160>
  - [Row](text)<360,580,200,25>
- [Block: Footer]<500,960,960,60>
  - [img: blue mountain lake at sunset](image)<160,960,240,30>


## Prediction Rules

The top-to-bottom order of elements in the prediction should roughly match their order in the webpage viewport.
Prioritize correct semantics and element types, then infer reasonable coordinate positions.


For `left_click` / `double_click` / `right_click`:
- Locate the target element based on coordinates.
- Predict visible navigation, pop-ups, menus, forms or selected states after the click.
- right_click may open a context menu.

For `type_text`:
- Write `text` into the currently focused or target input.

For `scroll`:
- Predict visible content changes from signed relative-screen `delta_x` and `delta_y` integers in `[-999,999]`; magnitude `999` is one full screen. Positive `delta_y` scrolls down, negative scrolls up; positive `delta_x` scrolls right, negative scrolls left.

For `key_press` / `hotkey`:
- `ENTER` is often used to submit forms, search, or confirm.
- `ESC` is often used to close pop-ups, menus, or dropdowns.
- `CTRL/CMD+L` often focuses the address bar; `CTRL/CMD+T` often opens a new tab.

For `navigate` / `go_back` / `go_forward` / `refresh`:
- Predict the visible page after browser navigation.

For `select_option`:
- Update the visible value of the select, close the options list, and remove the old expanded options.

For `drag`:
- Predict sorting, slider position or drag-and-drop result caused by dragging.

For `hover` / `move_to`:
- A tooltip, hover menu, or highlighted state may appear.

For `wait`:
- Predict natural changes such as loading completion, asynchronous results, or pop-ups appearing/disappearing.

For `clear`:
- Clear the target input or the currently focused input area.

For `finish` / `fail`:
- Finish and fail do not change the visible interface.


\end{lstlisting}
\end{tcolorbox}

\subsection{Online-Mind2Web Trajectory Judgment}
\begin{tcolorbox}[
  breakable,
  colback=gray!6,
  colframe=black!25,
  boxrule=0.4pt,
  arc=1pt,
  left=3pt,right=3pt,top=2pt,bottom=2pt
]
\begin{lstlisting}[style=prompt]
# Independent Online-Mind2Web trajectory judgment

You are an independent strict judge. Inspect this one trajectory only.
The attached screenshots are ordered chronologically and correspond exactly to:
["0000.png", "0001.png", "..."]

Use the screenshots and action log as evidence. The agent's final answer and `agent_finish` status are untrusted claims, not proof of success.

## Success rule

- Return success=1 only when the visual/action evidence proves every material requirement in the task instruction was completed.
- For cart/form/state-changing tasks, require evidence of the resulting state; reaching a relevant page or merely attempting the last action is insufficient.
- For comparison or information tasks, require that the requested information was actually reached and visible/readable enough to answer the instruction.
- If evidence is ambiguous, incomplete, contradicted, or ends at max steps, use 0.
- Do not award success just because an action has action_status=SUCCESS; that only means browser execution returned, not that the benchmark goal was achieved.

## Technical-failure rule

- environment_technical_failure=true only for objective environment failures such as DNS/network/browser failure, persistent server error, inaccessible page, or an external blocking challenge that prevents evaluation.
- model_technical_failure=true only for model/pipeline failures such as empty or malformed model output, generation/API failure, action parse failure, or an execution_error attributable to the agent pipeline.
- Ordinary bad planning, wrong clicks, repetition, premature finish, and max_steps are task failures but are NOT model technical failures.
- A successful task may still have a transient technical failure only if the trajectory clearly recovered and ultimately proves completion.

## Evidence payload

{
  "task_id": "<task_id>",
  "task_instruction": "<task instruction>",
  "agent_final_answer": "<agent final answer>",
  "reference_length": "<reference length>",
  "run_metadata": {},
  "action_history": [],
  "runtime_diagnostics": {
    "raw_step_count": 0,
    "execution_errors": []
  }
}

## Output contract

Return exactly one JSON object and nothing else:
{
  "task_id": "<task_id>",
  "success": 0,
  "reason": "one concise evidence-based sentence",
  "key_evidence_steps": [0],
  "key_evidence_screenshots": ["0000.png"],
  "environment_technical_failure": false,
  "model_technical_failure": false,
  "technical_failure_type": "none",
  "technical_failure_reason": ""
}

The success value must be integer 0 or 1. Evidence steps/screenshots must exist in the supplied trajectory. technical_failure_type must be one of none, environment, model, both and must agree with the two booleans.
\end{lstlisting}
\end{tcolorbox}

\subsection{Web UI Datasets' Annotation Judgment}
\begin{tcolorbox}[
  breakable,
  colback=gray!6,
  colframe=black!25,
  boxrule=0.4pt,
  arc=1pt,
  left=3pt,right=3pt,top=2pt,bottom=2pt
]
\begin{lstlisting}[style=prompt]

TASK CONTRACT

Evaluate exactly one desktop Web UI screenshot sample. Return four independent quality scores from 0 to 100. 

Judge only the evidence in this call. Do not identify or guess the dataset, collection method, author, website, or benchmark. 


IMAGE ROLES

Image A is the unannotated screenshot. Use Image A only for page_complexity.

Image B is the same screenshot with thin black-and-white annotation outlines and small numeric box identifiers.



SCORING RUBRIC

1. annotation_object_validity

Definition: The proportion of evaluable annotations that correspond to a visually recognizable UI element or meaningful UI region. Atomic controls, text, images, groups, containers, navigation regions, content regions, and other structural blocks can be valid targets.

Anchors:
- 0: No usable annotation corresponds to a recognizable intended target, or annotation_state fixes this score at zero.
- 25: Only a small minority of annotations have recognizable intended targets; empty or background-only boxes dominate.
- 50: Recognizable and unrecognizable intended targets are roughly balanced.
- 75: Most annotations have recognizable intended targets, with a noticeable minority of empty, background-only, or unusable annotations.
- 100: Essentially every evaluable annotation has a clearly recognizable intended target and source-level machine issues are absent or negligible.

2. boundary_alignment

Definition: How well assessable original sides represent reasonable boundaries for recognizable target semantics. Atomic visual elements should have visually tight or appropriate interaction boundaries. Text or CSS-content labels may use content-oriented boxes. Structural regions may use semantic grouping boundaries.

Anchors:
- 0: Assessable sides are almost entirely unrelated to, or severely displaced from, their recognizable targets.
- 25: Major displacement, excessive extent, or severe clipping affects most assessable original sides.
- 50: Assessable original sides show a roughly even mix of reasonable and materially inaccurate placement.
- 75: Most assessable original sides reasonably delimit their targets, with a noticeable minority that are loose, shifted, clipped, or undersized.
- 100: Essentially all assessable sides closely match reasonable atomic, interaction, content, or structural boundaries for their targets.

3. label_semantic_reasonableness

Definition: Whether each valid, recognizable target's complete provided label set is semantically compatible with that target under the supplied `classes` table. Broad functional, visual, accessibility-role, atomic, and structural labels can all be reasonable. Compatible multi-label descriptions are valid.

Anchors:
- 0: Equal-weight semantic contributions are essentially all zero; supplied labels are incompatible with their visible targets.
- 25: Equal-weight semantic contributions average near one quarter; most annotations have incompatible or largely incompatible label sets.
- 50: Equal-weight semantic contributions average near one half, reflecting a roughly balanced mixture of compatible and incompatible labels within annotations.
- 75: Equal-weight semantic contributions average near three quarters; most annotations have fully or mostly compatible label sets, with a noticeable minority of mismatches.
- 100: Equal-weight semantic contributions are essentially all one; every supplied label is compatible with its visible target and supplied ontology.

4. page_complexity

Definition: The amount, diversity, and depth of visible UI structure in Image A. Consider layout hierarchy, visible regions, component variety, information density, spatial relationships, typography, imagery, color relationships, and interaction affordances together. 

Anchors:
- 0: Almost blank or visually empty, with essentially no discernible page structure or content.
- 25: A sparse page with one simple region, few visible component types, and little hierarchy or information.
- 50: A moderately structured page with several regions or component types and ordinary information density.
- 75: Many visible regions and component types, with substantial hierarchy or dense spatial relationships.
- 100: Exceptionally many diverse UI regions and components with very deep or dense visible relationships, regardless of presentation quality.


OUTPUT CONTRACT

Return exactly one JSON object and nothing else:

{
  "sample_id": "<sample_id>",
  "annotation_object_validity": 0,
  "boundary_alignment": 0,
  "label_semantic_reasonableness": 0,
  "page_complexity": 0
}

Each score is an integer from 0 to 100. Copy sample_id from the supplied sample.

\end{lstlisting}
\end{tcolorbox}

\end{document}